\documentclass[a4paper,fleqn]{cas-sc}

\usepackage[numbers,sort&compress]{natbib}

\usepackage{graphicx}%
\usepackage{multirow}%
\usepackage{amsmath,amssymb,amsfonts}%
\usepackage{amsthm}%
\usepackage{mathrsfs}%
\usepackage[title]{appendix}%
\usepackage{xcolor}%
\usepackage{textcomp}%
\usepackage{manyfoot}%
\usepackage{booktabs}%
\usepackage{algorithm}%
\usepackage{algorithmicx}%
\usepackage{algpseudocode}%
\usepackage{listings}%
\usepackage{hyperref}
\usepackage{framed}
\usepackage{bm}
\usepackage{tabularx}
\usepackage{caption}
\usepackage{multirow}
\usepackage{booktabs}
\usepackage{amssymb}

\def\tsc#1{\csdef{#1}{\textsc{\lowercase{#1}}\xspace}}
\tsc{WGM}
\tsc{QE}

\begin{document}
\let\WriteBookmarks\relax
\def\floatpagepagefraction{1}
\def\textpagefraction{.001}

\shorttitle{}    

\shortauthors{}  

\title [mode = title]{YOLOv11-Litchi: Efficient Litchi Fruit Detection based on UAV-Captured Agricultural Imagery in Complex Orchard Environments}  



%
\author[1,2]{Hongxing Peng}[orcid=0000-0002-1872-8855]


\ead{xyphx@scau.edu.cn}




\author[1]{Haopei Xie}
\fnmark[1]

\ead{15907678645@163.com}


\author[1]{Weijia Li}
\fnmark[1]
\ead{18475888920@163.com}

\author[3]{Huanai Liu}

\ead{liuhn@scut.edu.cn}

\author[1]{Ximing Li}

\cormark[1]


\ead{liximing@scau.edu.cn}



\fntext[1]{These authors contributed equally to this work.}
\cortext[1]{Corresponding author}


\affiliation[1]{organization={South China Agricultural University},
            addressline={College of Mathematics and Informatics}, 
            city={Guangzhou},
            postcode={510642}, 
            country={China}}

\affiliation[2]{organization={Ministry of Agriculture and Rural Affairs},
            addressline={Key Laboratory of Smart Agricultural Technology in Tropical South China}, 
            city={Guangzhou},
            postcode={510642}, 
            country={China}}

\affiliation[3]{organization={South China University of Technology},
            addressline={School of Chemistry and Chemical Engineering}, 
            city={Guangzhou},
            postcode={510641}, 
            country={China}}

\begin{abstract}
Litchi is a high-value fruit, yet traditional manual selection methods are increasingly inadequate for modern production demands. Integrating UAV-based aerial imagery with deep learning offers a promising solution to enhance efficiency and reduce costs. This paper introduces YOLOv11-Litchi, a lightweight and robust detection model specifically designed for UAV-based litchi detection. Built upon the YOLOv11 framework, the proposed model addresses key challenges such as small target size, large model parameters hindering deployment, and frequent target occlusion. To tackle these issues, three major innovations are incorporated: a multi-scale residual module to improve contextual feature extraction across scales, a lightweight feature fusion method to reduce model size and computational costs while maintaining high accuracy, and a litchi occlusion detection head to mitigate occlusion effects by emphasizing target regions and suppressing background interference. Experimental results validate the model's effectiveness. YOLOv11-Litchi achieves a parameter size of 6.35 MB—32.5\% smaller than the YOLOv11 baseline—while improving mAP by 2.5\% to 90.1\% and F1-Score by 1.4\% to 85.5\%. Additionally, the model achieves a frame rate of 57.2 FPS, meeting real-time detection requirements. These findings demonstrate the suitability of YOLOv11-Litchi for UAV-based litchi detection in complex orchard environments, showcasing its potential for broader applications in precision agriculture.
\end{abstract}




\maketitle

\section{Introduction}\label{sec1}
Litchi is a tropical fruit widely cultivated in China and Southeast Asia. Its delicious taste and high economic value make it popular among consumers \cite{kuang2023residue}. Due to its high market value, both fresh litchi and its by-products are produced on a large scale in many countries, with China being the world's largest litchi producer \cite{qi2022method}. As the variety and planting area of litchi continue to expand, productivity has steadily increased. However, traditional manual methods of litchi fruit selection face significant challenges, including being time-consuming, imprecise, and costly, making them insufficient to meet modern production needs.

With the transition from traditional to intelligent agriculture \cite{yan2011design} and precision agriculture \cite{auernhammer2001precision}, deep learning technology has emerged as a powerful tool to accurately acquire agricultural production information and provide enhanced decision support. Applying deep learning to litchi detection tasks can effectively overcome the limitations of traditional manual methods by offering faster, more accurate, and cost-efficient solutions.

Nevertheless, the effective implementation of deep learning for litchi detection requires high-quality data. Litchi orchards often span large areas with rugged terrain, and litchi fruits are densely distributed at various angles, making comprehensive data collection challenging. To address these difficulties, unmanned aerial vehicle (UAV) remote sensing is an efficient alternative for capturing litchi fruit images. Compared with ground-based machinery and manual photography, UAVs can navigate rugged orchard terrain and capture high-quality aerial images \cite{alwateer2019enabling}. Due to their simplicity and ease of use, agricultural UAVs have been widely adopted in various fields, including plant protection \cite{cui2024weed,gao2024cross}, crop monitoring \cite{lee2023single,azizi2024comprehensive}, yield estimation \cite{liang2024rotated}, and pest detection \cite{joshi2024detection,tetila2020detection}.

While UAVs combined with deep learning have been successfully applied to crop detection tasks, such as longan \cite{li2021fast} and rapeseed \cite{li2024unmanned}, UAV-based litchi detection faces specific challenges. One major issue is occlusion: litchi fruits often grow in clusters, leading to overlapping among fruits. Additionally, leaves and branches can obstruct litchi fruits in aerial images, blurring target boundaries and disrupting feature structures. These occlusions increase the likelihood of missed or false detections. Furthermore, at higher UAV flight altitudes, litchi fruits appear smaller in the images, which poses additional difficulties for accurate detection.

Real-time litchi detection using UAVs also imposes strict requirements on the efficiency and size of the detection model. Although advanced methods like Transformer \cite{ashish2017attention} and DETR \cite{carion2020end} achieve high detection accuracy, they often demand significant computational resources. Given the resource constraints typical in UAV deployment scenarios, these methods are not always practical for real-world applications.

To address these challenges, this paper proposes a novel litchi detection algorithm tailored for UAV-based applications in complex orchard environments. The main contributions of this paper are as follows:
\begin{enumerate}
    \item To address the issue of small litchi targets in UAV imagery, we improve the C3 module and propose the C3 multi-scale residual(C3-MSR) module, enhancing the model's capability to extract multi-scale features without increasing computational overhead. This improves the model's performance in detecting litchi fruits in UAV scenarios.
    \item To meet the real-time detection requirements of UAVs, we design a lightweight feature fusion method that reduces model parameters and computational costs while maintaining detection accuracy. This enables the model to operate effectively in resource-constrained environments.
    \item To handle occlusion issues, we classify occlusions into three categories: no occlusion, fruit occlusion, and branch or leaf occlusion. Using the Self-Ensembling Attention Mechanism(SEAM), we design the SEAM-Head module to enhance the model's ability to learn litchi features, reducing the missed detection rate under occlusion conditions.
    \item The proposed algorithm achieves a model parameter size of 6.35 MB, which is 32.5\% smaller than the YOLOv11 benchmark network, while improving mAP by 2.5\% to reach 90.1\%. The model achieves the best balance between accuracy and speed, and generalization experiments further validate its robustness and applicability to other crop detection tasks.
\end{enumerate}

\section{Related Work}\label{sec2}

\subsection{Occlusion Problem}

Occlusion is a common challenge in complex scenes and is one of the primary factors leading to decreased target detection accuracy. For example, researchers \cite{gene2020fruit} applied the Mask R-CNN \cite{he2017mask} neural network to detect apples but found that heavily occluded fruits were difficult to identify. Current technologies often overlook the issue of severe fruit overlap. However, litchi fruits, which typically grow in clusters, are particularly prone to occlusion caused by overlapping fruits. This results in blurred or invisible boundaries in certain areas, leading to missed detections and low recall rates.

To address the occlusion problem, various methods have been proposed. SSH \cite{najibi2017ssh} employs a simple convolutional layer to aggregate contextual information by expanding the region of interest around the target, thereby improving the extraction of valuable information from occluded areas. FAN \cite{wang2017face} introduces an anchor-level attention mechanism that highlights key features in occluded regions to enhance detection performance. Other studies \cite{wang2021diseases} suggest expanding feature map channels to extract high-dimensional features and then reducing dimensionality to improve the algorithm's capacity for occluded target detection. DSW-YOLO \cite{du2023dsw} addresses the occlusion problem by enhancing the network's ability to extract features from unconventional targets during strawberry fruit detection. Similarly, an active depth perception method \cite{sun2024efficient} has been proposed to harvest both clusters and individual fruits by leveraging neural networks to identify regions of interest and employing image processing to assess occlusion states.

These methods have demonstrated promising results in addressing occlusion challenges. However, no comprehensive solution exists for handling litchi occlusion from the UAV perspective. To address this gap, this paper proposes a litchi occlusion detection head based on an occlusion attention mechanism. Building on the aforementioned methods, the proposed approach leverages contextual information to emphasize litchi regions while suppressing background areas. This ensures a stronger focus on critical features, thereby mitigating the impact of occlusion on detection accuracy.

\subsection{Multi-Scale Feature Fusion}

Efficiently representing and processing multi-scale features is a key challenge in target detection, as objects of different scales often exhibit distinctive and identifiable characteristics. Early detectors directly utilized features extracted from backbone networks for prediction \cite{cai2016unified,liu2016ssd}. The introduction of the feature pyramid network (FPN) \cite{lin2017feature} marked a significant milestone in addressing this challenge. FPN facilitates the fusion of multi-scale features through cross-scale connections and information exchange, achieving remarkable improvements in the detection accuracy of objects at varying scales.

Building on FPN, numerous cross-scale feature fusion network structures have been developed. For instance, PANet \cite{liu2018path} enhances information flow by incorporating a bottom-up pathway. EfficientDet \cite{tan2020efficientdet} introduced the bi-directional feature pyramid network (BiFPN), which utilizes learnable weights to balance the importance of input features and performs repeated top-down and bottom-up multi-scale feature fusion. Compared to FPN, BiFPN achieves more comprehensive utilization of multi-scale features. PRB-FPN \cite{chen2021parallel} proposed a parallel FPN structure that supports two-way feature fusion, addressing the diminishing effectiveness of FPN at deeper network levels. AFPN \cite{yang2023afpn} extends FPN by breaking its limitations in detecting large targets and enables cross-layer interactions between non-adjacent layers. Furthermore, Gold-YOLO \cite{wang2024gold} incorporates a global-local feature fusion strategy, effectively balancing global and local features to enhance multi-scale feature fusion capabilities.

Efficient multi-scale feature fusion is critical for litchi detection tasks in UAV scenarios, as litchi fruit size in images varies significantly depending on UAV altitude and camera angle. Building on the aforementioned studies, this paper adopts a multi-scale feature fusion approach to improve the detection accuracy of litchi fruits in UAV images.

\section{Materials and methods}\label{sec3}
\subsection{Data Collection}

\begin{figure}
    \centering
    \includegraphics[width=1\linewidth]{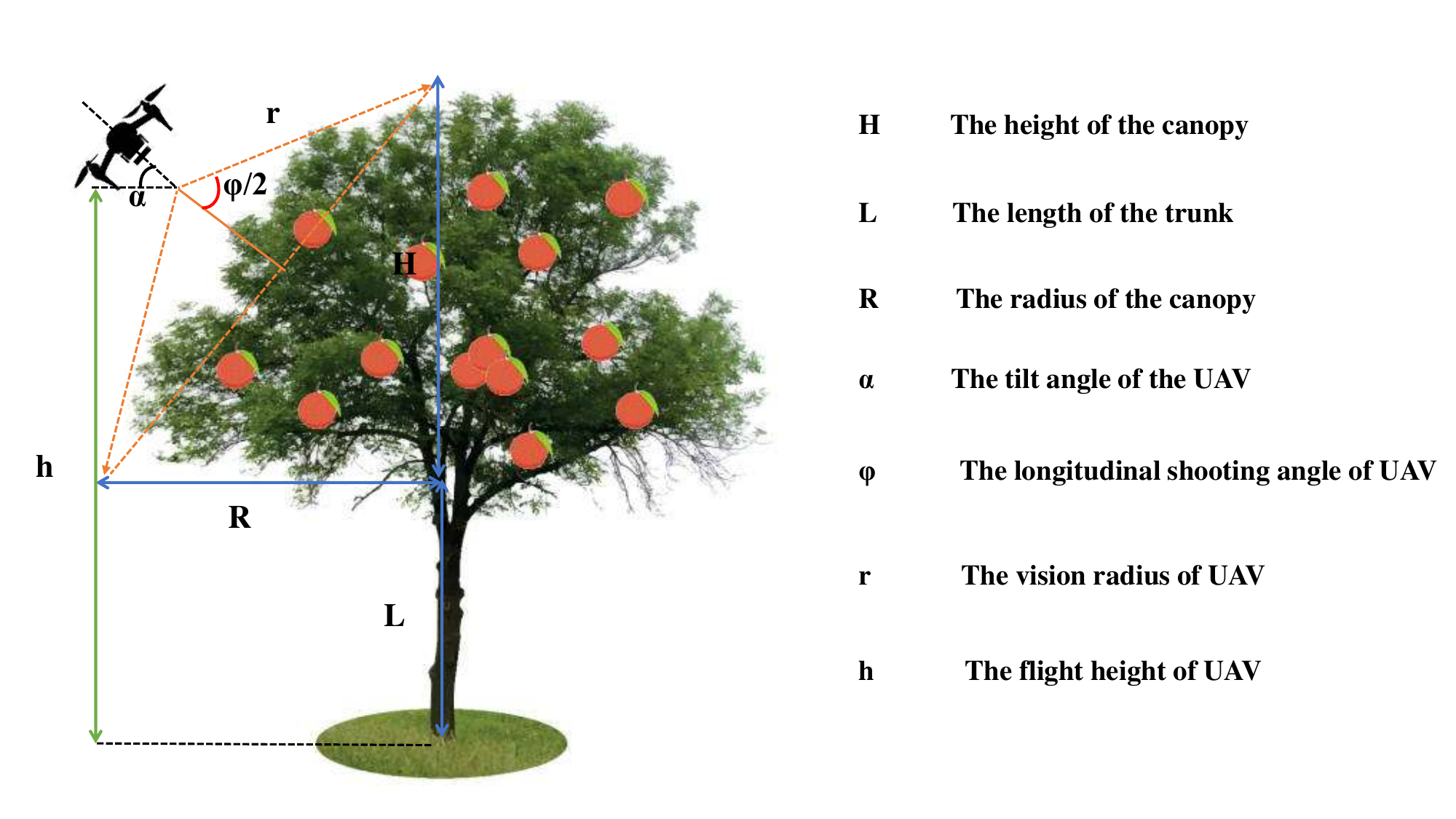}
    \caption{Tilt shooting method of the UAV.}
    \label{fig:UAV}
\end{figure}

\begin{figure}
    \centering
    \includegraphics[width=0.95\linewidth]{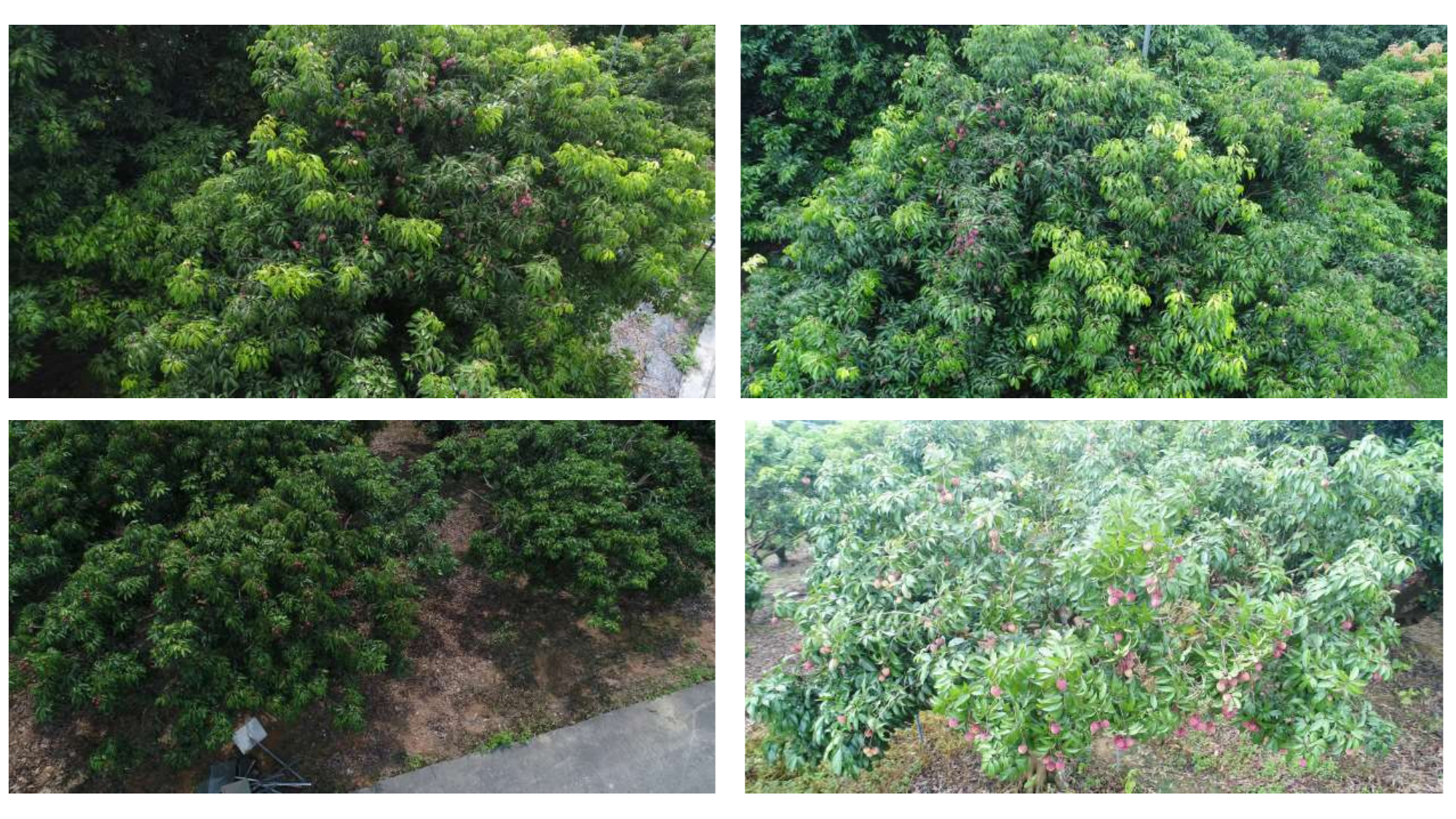}
    \caption{Example of litchi images captured by the UAV.}
    \label{fig:litchi}
\end{figure}

\begin{figure}
    \centering
    \includegraphics[width=0.95\linewidth]{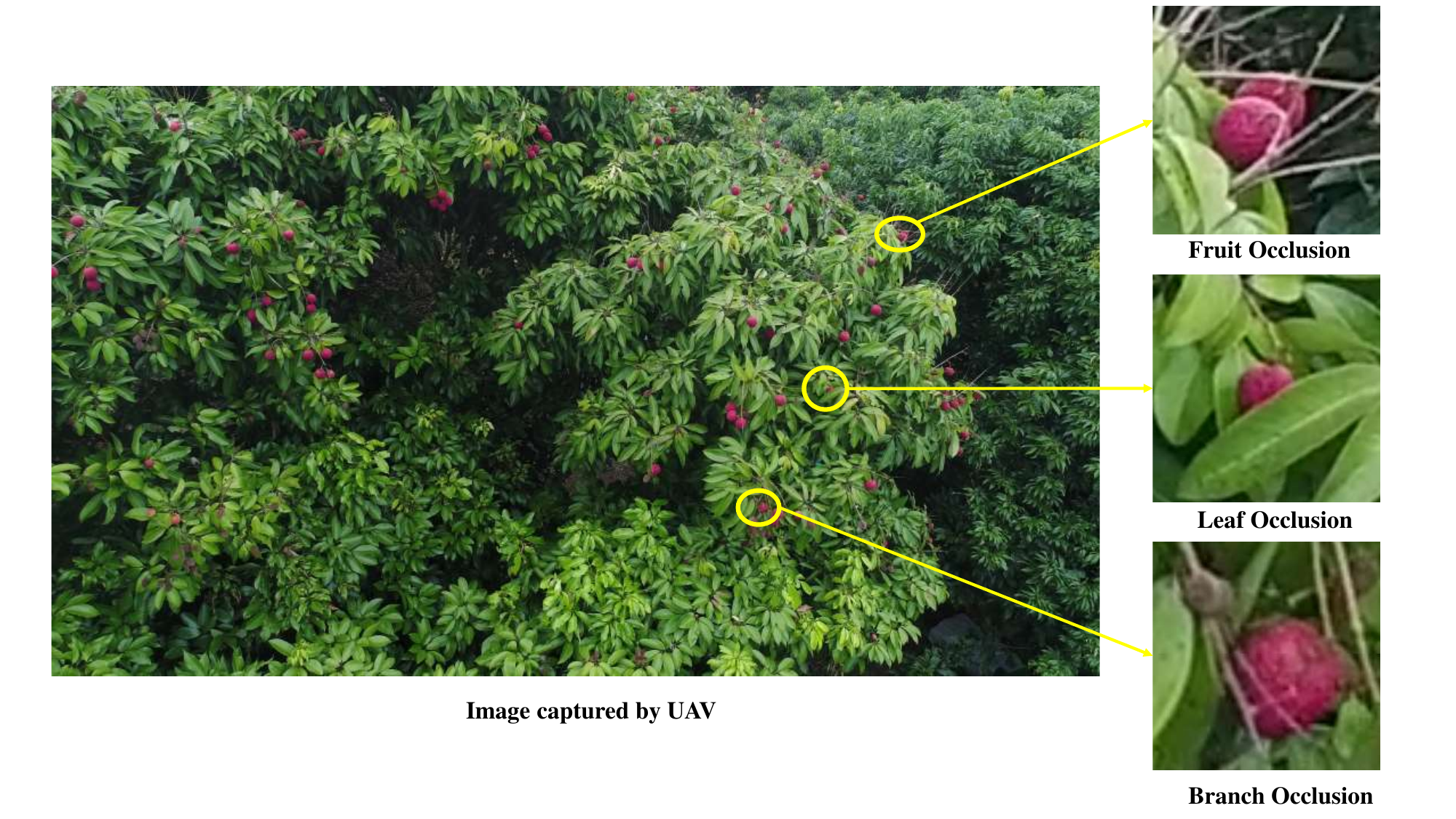}
    \caption{Different occlusion types observed in litchi images.}
    \label{fig:occlusion}
\end{figure}

The image data for this study were collected on May 14th, 2024 (cloudy to overcast), and July 2nd, 2024 (sunny), at the Litchi Culture Expo Park (113° 618' E, 23° 583' N) in Conghua District, Guangzhou. The images, with a resolution of 4096×2160 pixels, were captured using an UAV(DJI Elf 4). Two shooting methods were employed: vertical shooting and oblique shooting.

In the vertical shooting method, special attention was given to the strong downward airflow generated by the UAV. When the UAV flies too close to the litchi trees, this airflow can knock off branches or fruits, potentially causing economic losses. To mitigate this risk, the UAV's flight height was pre-determined based on field experiments. It was observed that when the UAV's flight height exceeded the fruit tree canopy by 3 meters or more, it did not affect the trees. Therefore, the UAV was set to fly at a height equal to the average canopy height of the fruit trees plus an additional 3–5 meters. Once the flight height was determined, the UAV followed a predefined path through the orchard, capturing images with its camera oriented vertically downward.

For the oblique shooting method, the UAV's angle of capture was adjusted to ensure coverage of the litchi tree canopy. As illustrated in Figure~\ref{fig:UAV}, parameters such as canopy height ($h$), canopy radius ($r$), trunk length ($l$), and the UAV's longitudinal shooting angle ($\phi$) were measured. These measurements were used to calculate the tilt angle ($\alpha$) of the UAV's lens, the vision radius ($r$), and the flight height ($h$) of the UAV using the following equations:

\begin{equation}
\alpha = \tan^{-1} \left( \frac{R}{H} \right)
\end{equation}

\begin{equation}
r = \frac{\sqrt{R^{2} + H^{2}}}{2 \sin \left( \frac{\varphi}{2} \right)} \times \cos \left( \frac{\varphi}{2} - \alpha \right)
\end{equation}

\begin{equation}
h = \frac{\sqrt{R^{2} + H^{2}}}{2 \sin \left( \frac{\varphi}{2} \right)} \times \sin \left( \frac{\varphi}{2} + \alpha \right) + L
\end{equation}

After eliminating duplicate, low-quality, and excessively dark images, a total of 432 original images were retained. An example of such images are shown in Figure~\ref{fig:litchi}. And the litchi images exhibit various types of occlusions, including fruit occlusion, branch occlusion, and leaf occlusion, as illustrated in Figure~\ref{fig:occlusion}.

\subsection{Data Pre-processing}

\begin{figure}
    \centering
    \includegraphics[width=0.95\linewidth]{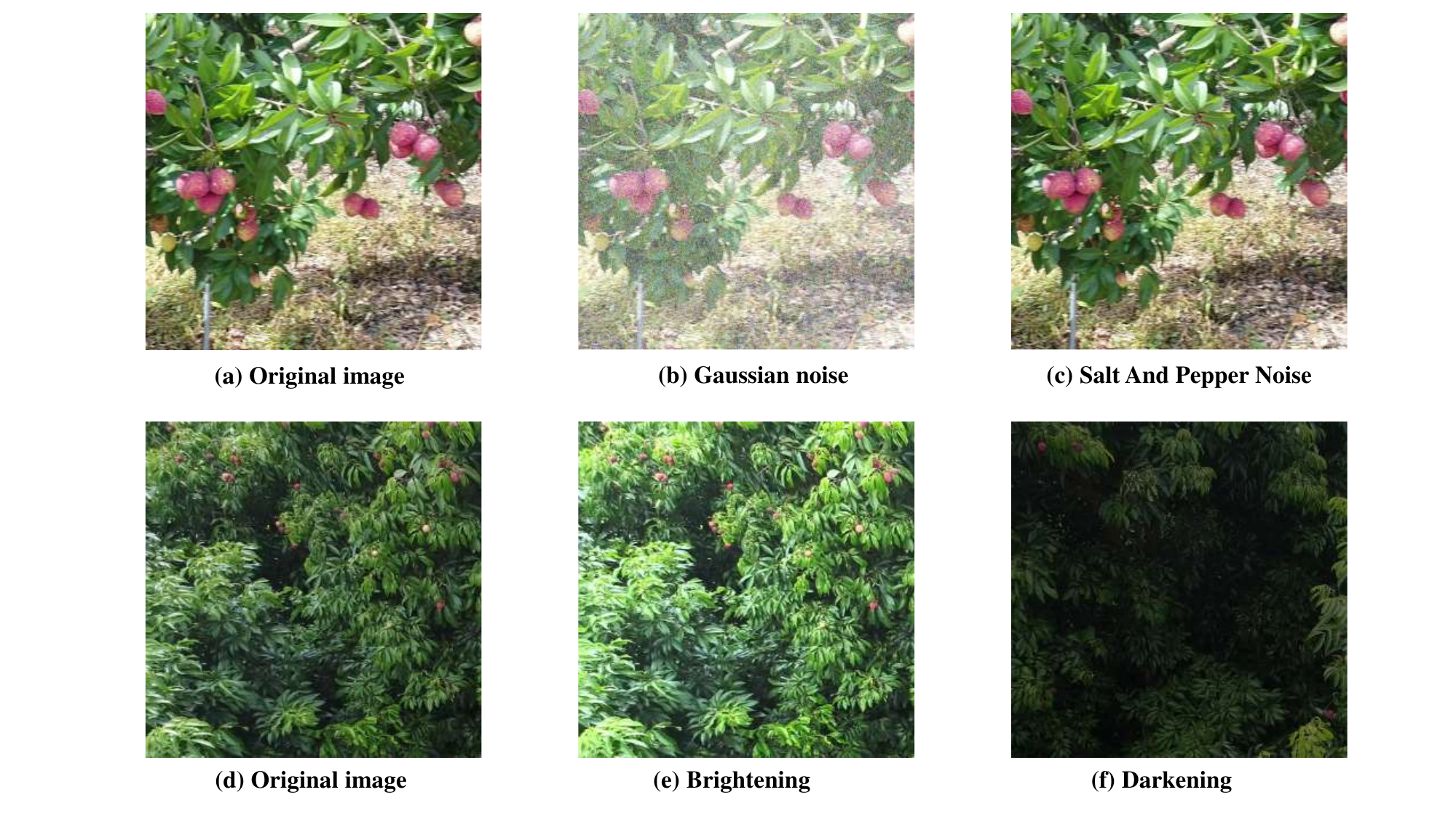}
    \caption{Image enhancement methods applied to litchi images.}
    \label{fig:augment}
\end{figure}

To prepare the data for training, the original images were divided into smaller segments using a sliding window of size 1024×1024. The resulting image patches were then split into three subsets: training set, validation set, and test set, in a ratio of 7:2:1.

To enhance the diversity of the training data and improve model robustness, four image augmentation strategies were applied: Gaussian noise, salt-and-pepper noise, image brightening, and image darkening. The effects of these augmentations are shown in Figure~\ref{fig:augment} (b), (c), (e), and (f), respectively. 

After augmentation, the dataset consisted of 849 images for training, 73 images for validation, and 56 images for testing, totaling 978 images. To distinguish this dataset from others used in subsequent experiments, it was named Litchi-UAV.

\subsection{Public Dataset}

\begin{figure}
    \centering
    \includegraphics[width=0.95\linewidth]{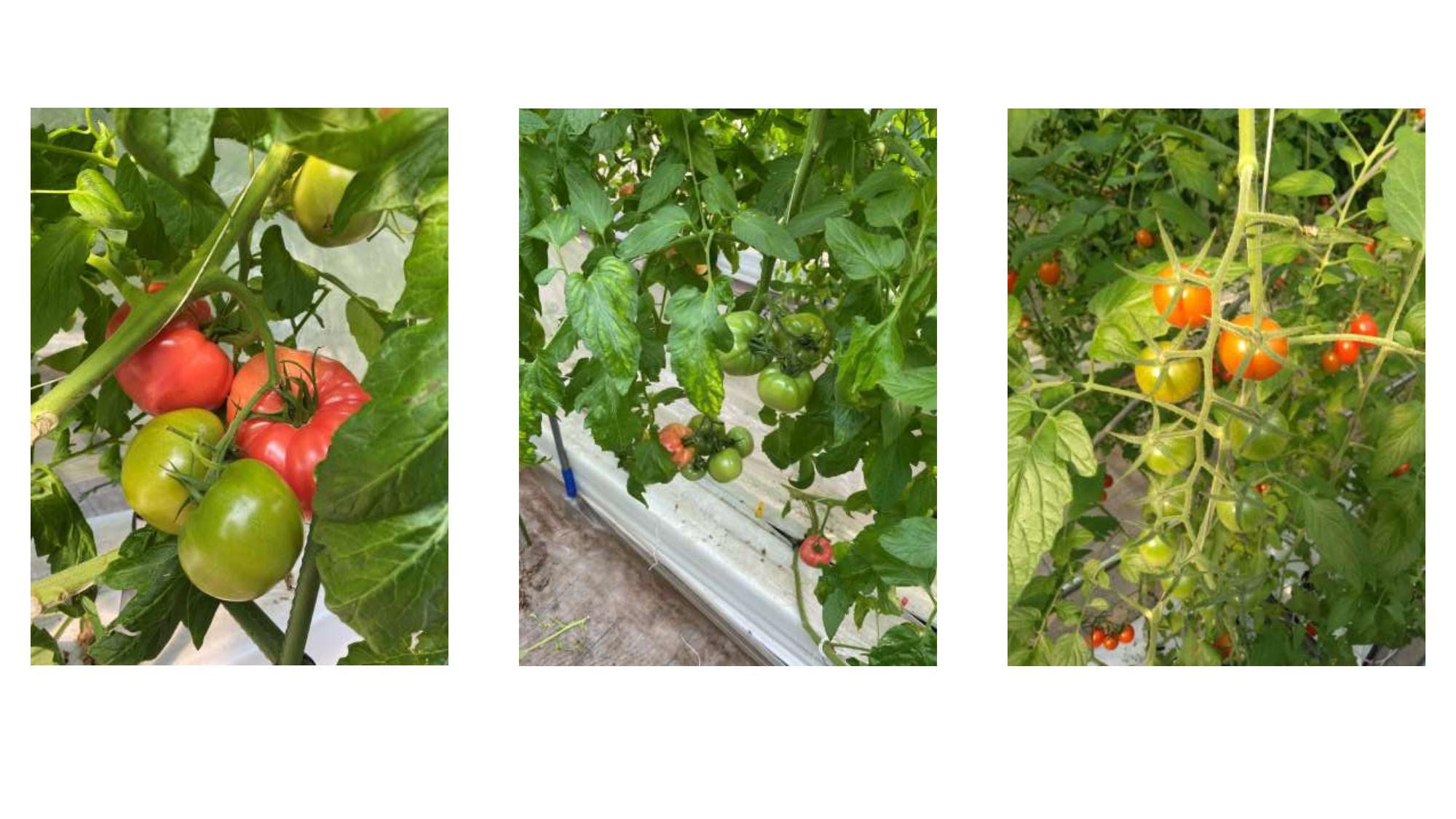}
    \caption{Sample image from the Laboro Tomato public dataset.}
    \label{fig:tomato}
\end{figure}

\begin{figure}
    \centering
    \includegraphics[width=0.95\linewidth]{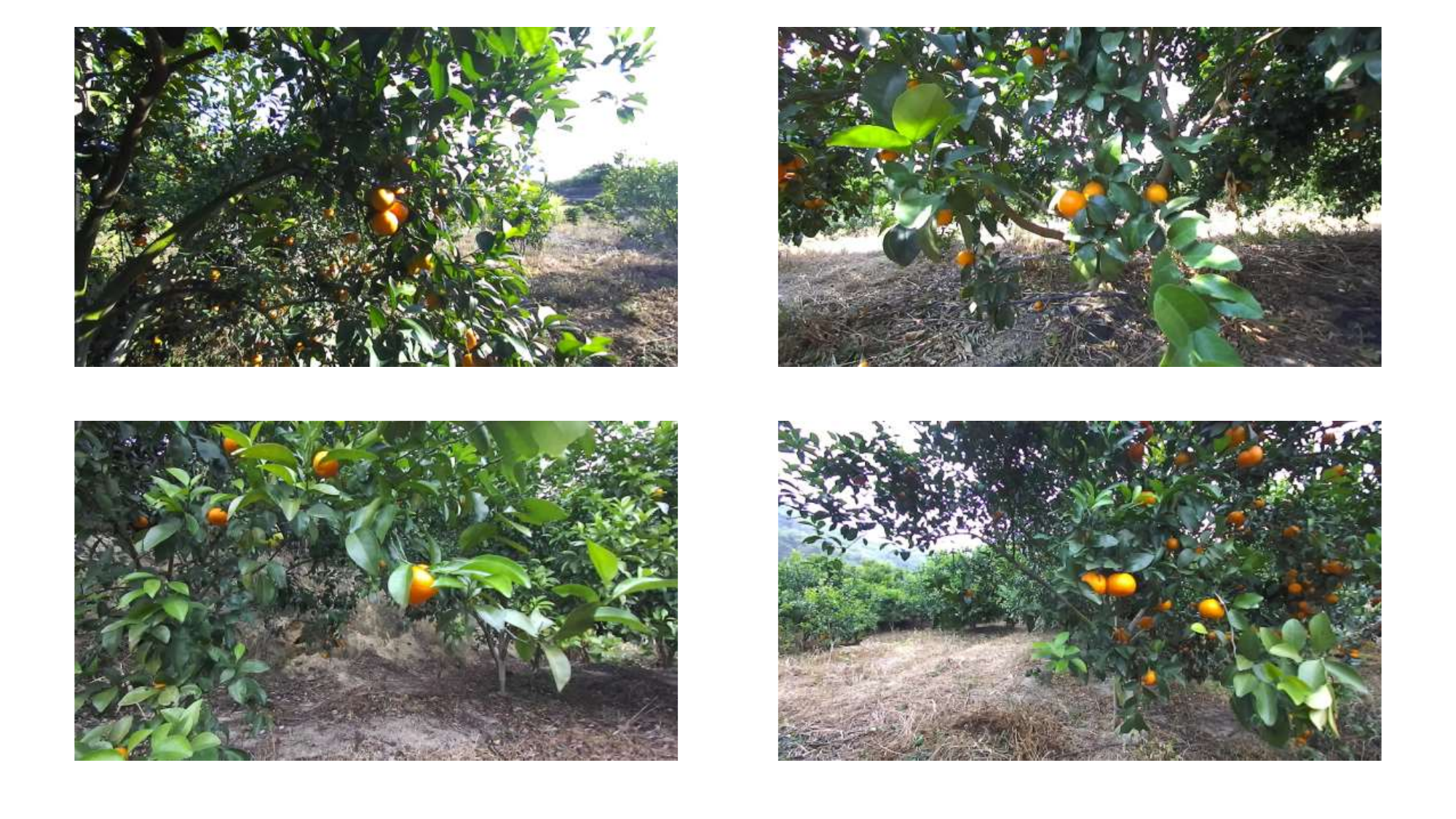}
    \caption{Sample image from the Citrus public dataset.}
    \label{fig:citrus}
\end{figure}

To evaluate the generalization ability of our proposed model on other datasets, we introduced two publicly available crop datasets: the Laboro Tomato Dataset\cite{laboro_tomato} and the Citrus Dataset\cite{hou2022detection}. These datasets were selected for their diversity in crop types and imaging conditions, providing a robust testbed for generalization experiments.

The Laboro Tomato Dataset is a public collection of tomato images capturing tomatoes at various stages of maturity. It is specifically designed for object detection and instance segmentation tasks. The dataset was acquired using two independent cameras with varying resolutions and image qualities. It contains a total of 804 images, subdivided into 643 training images and 161 testing images, encompassing approximately 10,000 labeled tomato instances. A sample image from this dataset is shown in Figure~\ref{fig:tomato}.

The Citrus Dataset is a public dataset of citrus images collected from hillside orchards at Conghua Hualong Fruit and Vegetable Fresh Co., Ltd., Guangzhou, China (113°39' E, 23°33' N). It consists of 4855 images captured at distances ranging from 30 to 150 cm between the camera and the citrus. Based on surface illumination conditions, the images are categorized into three groups: uneven illumination, weak illumination, and good illumination. The dataset is divided into 2913 training images, 971 validation images, and 971 testing images. A sample image from this dataset is shown in Figure~\ref{fig:citrus}.

\subsection{Overall Architecture}

\begin{figure}
    \centering
    \includegraphics[width=0.95\linewidth]{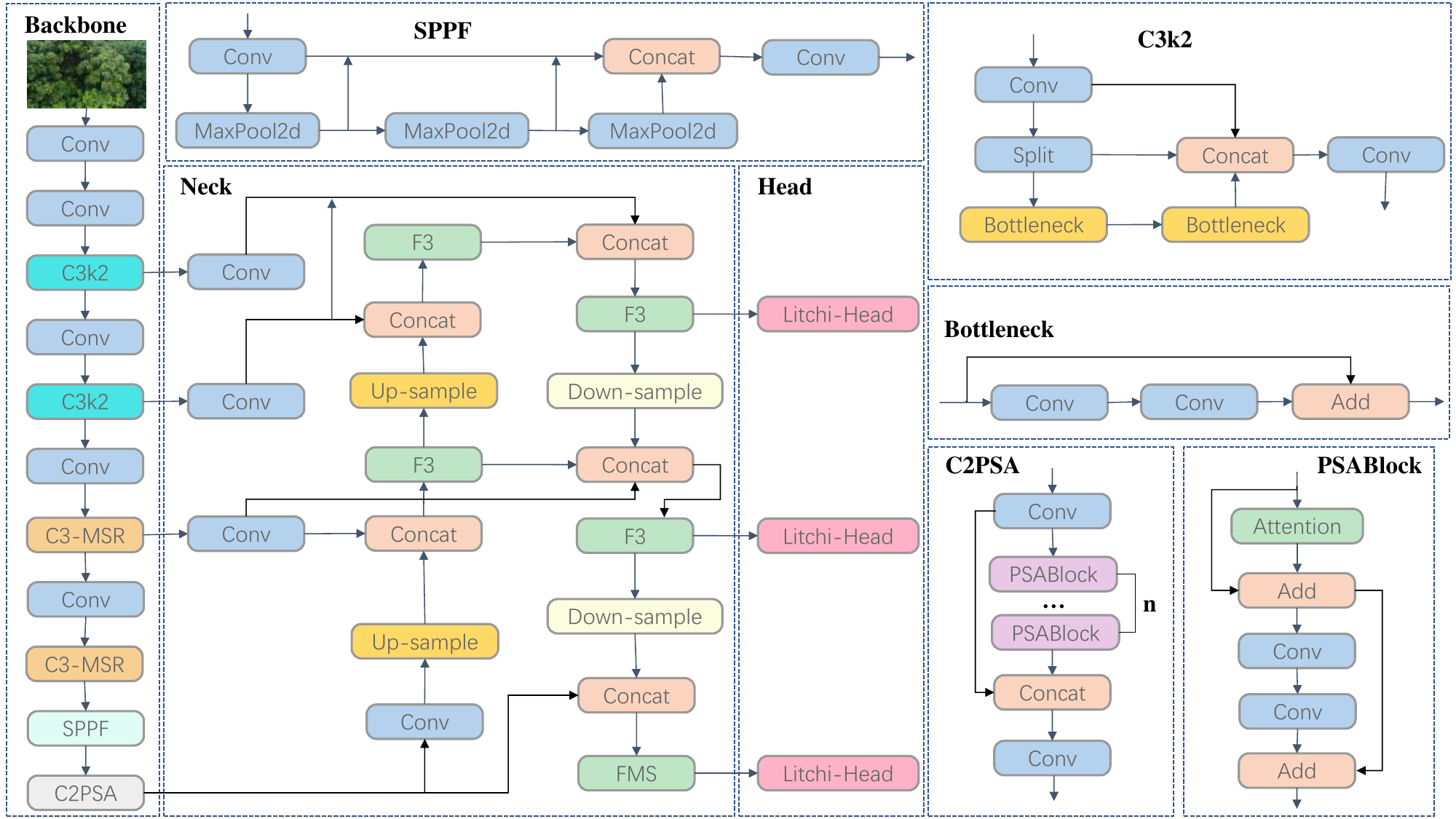}
    \caption{Overall framework of YOLOv11-Litchi.}
    \label{fig:model}
\end{figure}

To address the challenges of litchi detection in UAV-captured images, this study introduces a novel detection model, YOLOv11-Litchi. The model incorporates three key strategies: the Multi-Scale Residual Module, a Lightweight Feature Fusion Method, and a Litchi Occlusion Detection Head. These enhancements are designed to improve detection accuracy while maintaining computational efficiency, making the model highly suitable for UAV deployment. The overall structure of YOLOv11-Litchi is specifically optimized for the unique requirements of litchi detection in UAV imagery. Figure~\ref{fig:model} illustrates the model's streamlined architecture, demonstrating its capability for precise and efficient litchi detection.

\subsection{Multi-Scale Residual Module}

\begin{figure}
    \centering
    \includegraphics[width=0.95\linewidth]{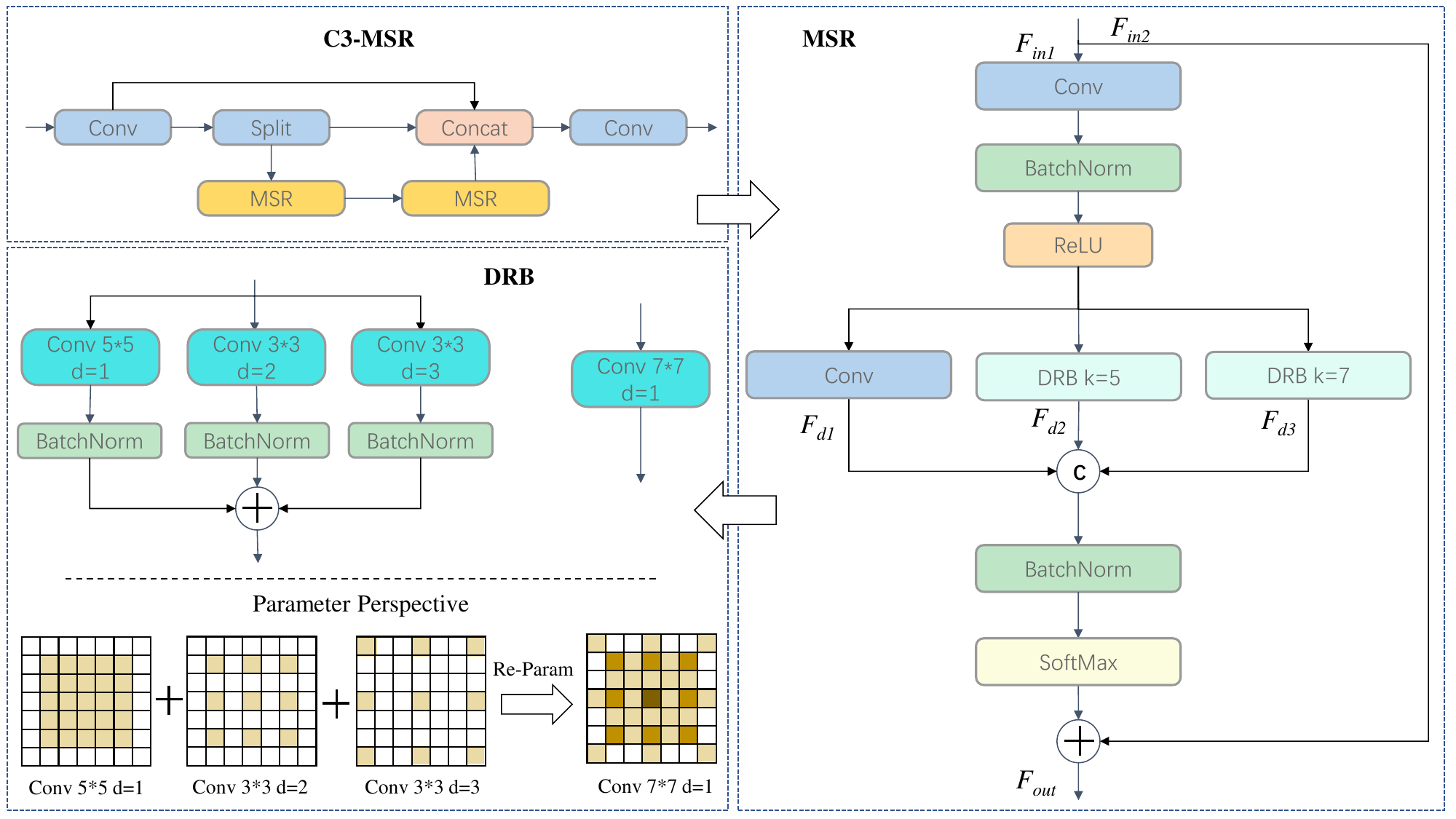}
    \caption{The details of C3-MSR.}
    \label{fig:backbone}
\end{figure}

In UAV imagery, the target object often occupies a very small portion of the overall field of view, placing high demands on the model's ability to extract contextual information at different scales during the detection process. This issue is particularly pronounced in the task of detecting litchi fruits in UAV-captured images, as litchi exhibits characteristics of small individual size and clustered growth. Consequently, efficiently and comprehensively extracting multi-scale features becomes crucial for UAV-based detection tasks. 

Conventional convolutional designs face inherent limitations in directly capturing multi-scale contextual information. This is primarily because traditional convolution operations process input data on a fixed spatial scale, with a predefined receptive field (i.e., the area of input data covered by the convolution kernel). As a result, such designs may struggle to effectively capture feature information across diverse scales. 

Inspired by DWRSeg\cite{wei2022dwrseg}, we optimized the C3 module structure by introducing the Multi-Scale Residual(MSR) module to replace the original Bottleneck component within the C3 module, yielding the C3-MSR module. This enhanced module can more efficiently extract multi-scale features, thereby improving the model's detection performance in UAV-based scenarios. The details of C3-MSR are showed in figure~\ref{fig:backbone}. And the key mathematical formulations for the C3-MSR module are presented below.

\begin{equation}
F_{\text{in1}}, F_{\text{in2}} = \text{Split}(F_{\text{in}})
\end{equation}
 
\begin{equation}
F_{d1} = \text{Conv}_{3 \times 3, d=1}(F_{\text{in1}})
\end{equation}
 
\begin{equation}
F_{d2} = \text{Concat}\left(\text{Conv}_{3 \times 3, d=1}(F_{\text{in1}}), \text{Conv}_{3 \times 3, d=2}(F_{\text{in1}})\right)
\end{equation}

\begin{equation}
F_{d3} = \text{Concat}\left(\text{Conv}_{5 \times 5, d=1}(F_{\text{in1}}), \text{Conv}_{3 \times 3, d=2}(F_{\text{in1}}), \text{Conv}_{3 \times 3, d=3}(F_{\text{in1}})\right)
\end{equation}
 
\begin{equation}
F_{\text{MSR}} = \text{Concat}\left(F_{d1}, F_{d2}, F_{d3}\right)
\end{equation}

\begin{equation}
F_{\text{out}} = \text{Add}(F_{\text{in2}}, F_{\text{MSR}})
\end{equation}

Our proposed design incorporates the residual learning concept from the ResNet\cite{he2016deep}, decomposing the typical one-step multi-scale context acquisition process into two branches: \(F_{\text{in1}}\) and \(F_{\text{in2}}\). The first branch, \(F_{\text{in1}}\), preserves the initial feature information, while the second branch, \(F_{\text{in2}}\), extracts multi-scale features through specialized multi-scale feature extraction mechanisms. The two branches are subsequently merged to produce a more comprehensive feature representation, \(F_{\text{out}}\).

For multi-scale feature extraction, we first apply a standard \(3 \times 3\) convolution for initial feature processing, followed by BatchNorm and ReLU layers for data normalization and activation. Subsequently, we extract features at varying scales using three dilated convolution branches with dilation rates of 1, 3, and 5. However, using large kernels and high dilation rates can significantly increase computational cost and introduce noise or redundant information, posing challenges for deployment on resource-constrained UAV platforms. To address this, inspired by UniRepLKNet\cite{ding2024unireplknet}, we replace large kernels in the branches with dilation rates of 3 and 5 using the Dilated Reparam Block(DRB). This approach employs re-parameterized smaller convolution kernels, achieving similar receptive field effects with reduced resource consumption.

Studies have shown that combining large kernel convolutions with parallel small kernel convolutions is beneficial, as the latter helps capture fine-grained features during training\cite{ding2022scaling}. Using re-parameterization techniques\cite{ding2021diverse}\cite{ding2021repvgg}\cite{ding2022re}, small kernels can emulate the functionality of larger kernels without incurring excessive computational overhead. This design allows the module to flexibly adapt to various input data types and task requirements.

The choice of convolution kernel structure significantly affects the model's feature extraction capabilities. In the first branch, we employ a single \(3 \times 3\) convolution to compute \(F_{d1}\). The second branch uses two \(3 \times 3\) convolutions, where the first has no dilation and the second has a dilation rate of 2. These are re-parameterized to produce \(F_{d2}\), focusing on local detail extraction. The third branch combines a \(5 \times 5\) convolution with two \(3 \times 3\) convolutions, applying dilation rates of 1, 2, and 3, respectively, to generate \(F_{d3}\). This branch captures richer contextual information due to its larger receptive field. These outputs, \(F_{d1}\), \(F_{d2}\), and \(F_{d3}\), are concatenated via the Concat operation to form \(F_{\text{MSR}}\). Finally, \(F_{\text{MSR}}\) is integrated with \(F_{\text{in2}}\) via a residual connection to produce the final output \(F_{\text{out}}\), yielding a robust and comprehensive feature representation. 

\subsection{Lightweight Feature Fusion Method}

\begin{figure}
    \centering
    \includegraphics[width=0.95\linewidth]{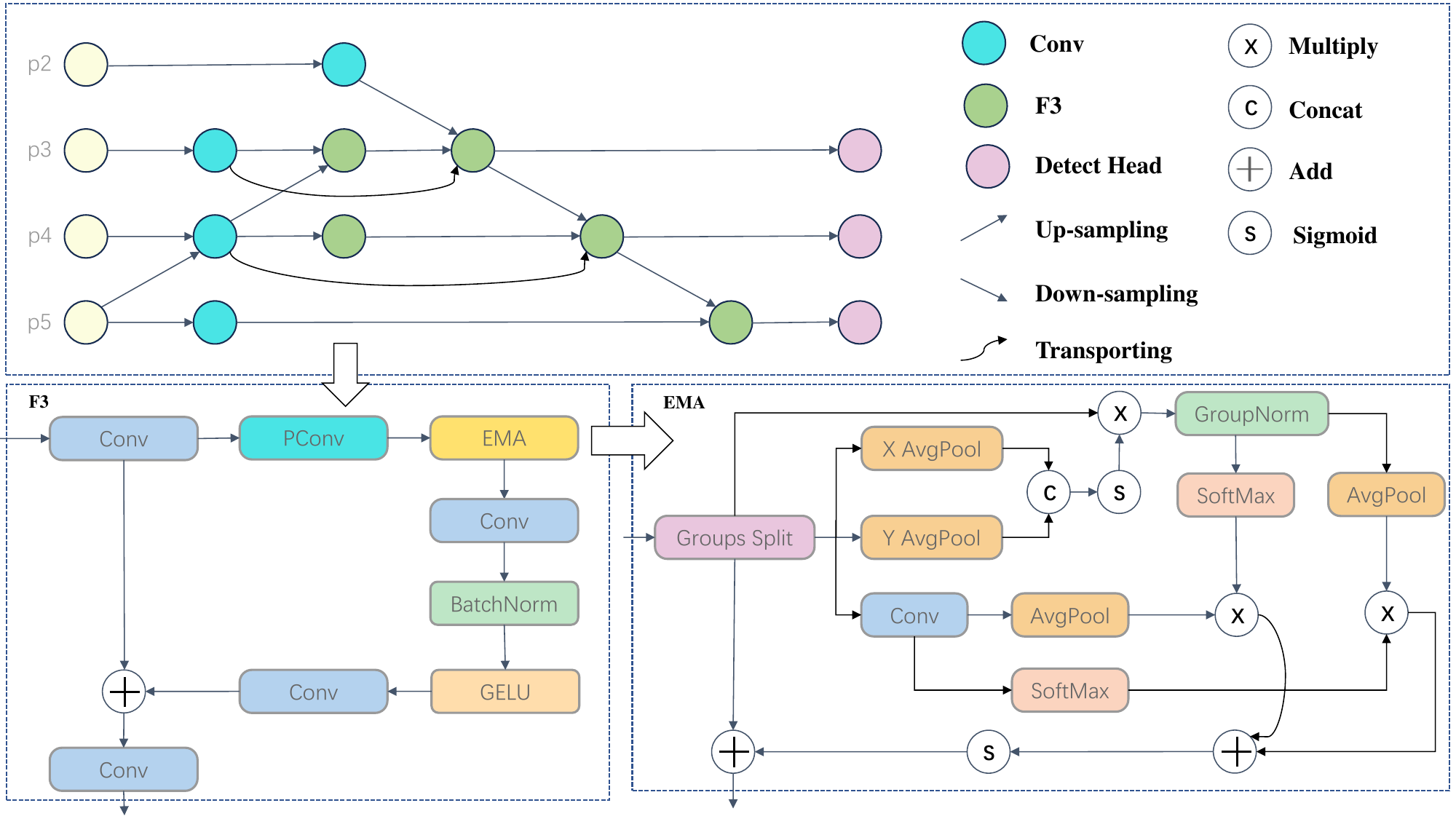}
    \caption{The details of lightweight feature fusion method.}
    \label{fig:neck}
\end{figure}

Many state-of-the-art(SOTA) detection models can achieve high-precision target detection, but they typically require significant computational resources, making them unsuitable for embedded hardware platforms such as UAVs. These limitations hinder their ability to meet the real-time detection requirements in UAV scenarios. To address this issue, we propose the Faster Feature Fusion(F3) module, inspired by the core principles of the weighted bi-directional feature pyramid network (BiFPN). This module is integrated into the neck of YOLOv11, enabling efficient and lightweight multi-scale feature fusion. The F3 module significantly reduces model parameters and computational demands while maintaining detection accuracy, thereby making the model suitable for real-time UAV applications. The mathematical formulation of the F3 module is as follows:

\begin{equation}
F_{c1}, F_{c2} = \text{Split}(\text{Conv}(F_{\text{in}}))
\end{equation}
 
\begin{equation}
F_{\text{ema}} = \text{EMA}(\text{PConv}(F_{c2}))
\end{equation}

\begin{equation}
F_{\text{out}} = \text{Add}(F_{c1}, F_{\text{ema}})
\end{equation}

In the F3 module, the input feature map is first processed by a \(3 \times 3\) convolution, which is then split into two branches: \(F_{c1}\) and \(F_{c2}\). The \(F_{c1}\) branch retains global features, while the \(F_{c2}\) branch focuses on extracting local features. The \(F_{c2}\) branch is further processed using PConv\cite{chen2023run}, which effectively reduces redundant computations and memory access while enhancing spatial feature extraction. 

Although PConv improves computational efficiency and reduces model parameters, it may lead to the loss of local feature fragments due to its compression process. To mitigate this issue, we introduce the Efficient Multi-Scale Attention Module (EMA)\cite{ouyang2023efficient} into the \(F_{c2}\) branch. EMA effectively preserves channel-wise information without introducing additional computational overhead. Specifically, this module learns an efficient channel representation without reducing channel dimensionality via convolution operations, enabling it to generate enhanced pixel-level attention for advanced feature maps.

The structure of EMA is depicted in the lower right of Figure \ref{fig:neck}. EMA reconstructs selected channels into batch dimensions and groups channel dimensions into multiple sub-features, ensuring the even distribution of spatial semantic features across each feature group. It then employs three parallel routes to extract attention weights for the grouped feature maps. To balance computational efficiency and cross-channel dependency modeling, EMA uses two two-dimensional global average pooling operations in two branches to encode global spatial channel information, while a \(3 \times 3\) convolution is applied in the third branch to capture multi-scale feature representations. These outputs are aggregated to generate a spatial attention map. The final feature map is derived by aggregating the output of each group, weighted by two Sigmoid functions for spatial attention, enabling the model to capture pixel-level relationships and highlight global pixel contexts. This design effectively addresses the problem of local feature loss.

Finally, the global features from \(F_{c1}\) and the enhanced local features from \(F_{\text{ema}}\) are fused to generate comprehensive image features. This fusion strategy simultaneously captures global feature control and retains detailed local feature learning. As a result, the F3 module achieves a lightweight model design without sacrificing detection accuracy.

\subsection{Litchi Occlusion Detection Head}

\begin{figure}
    \centering
    \includegraphics[width=0.95\linewidth]{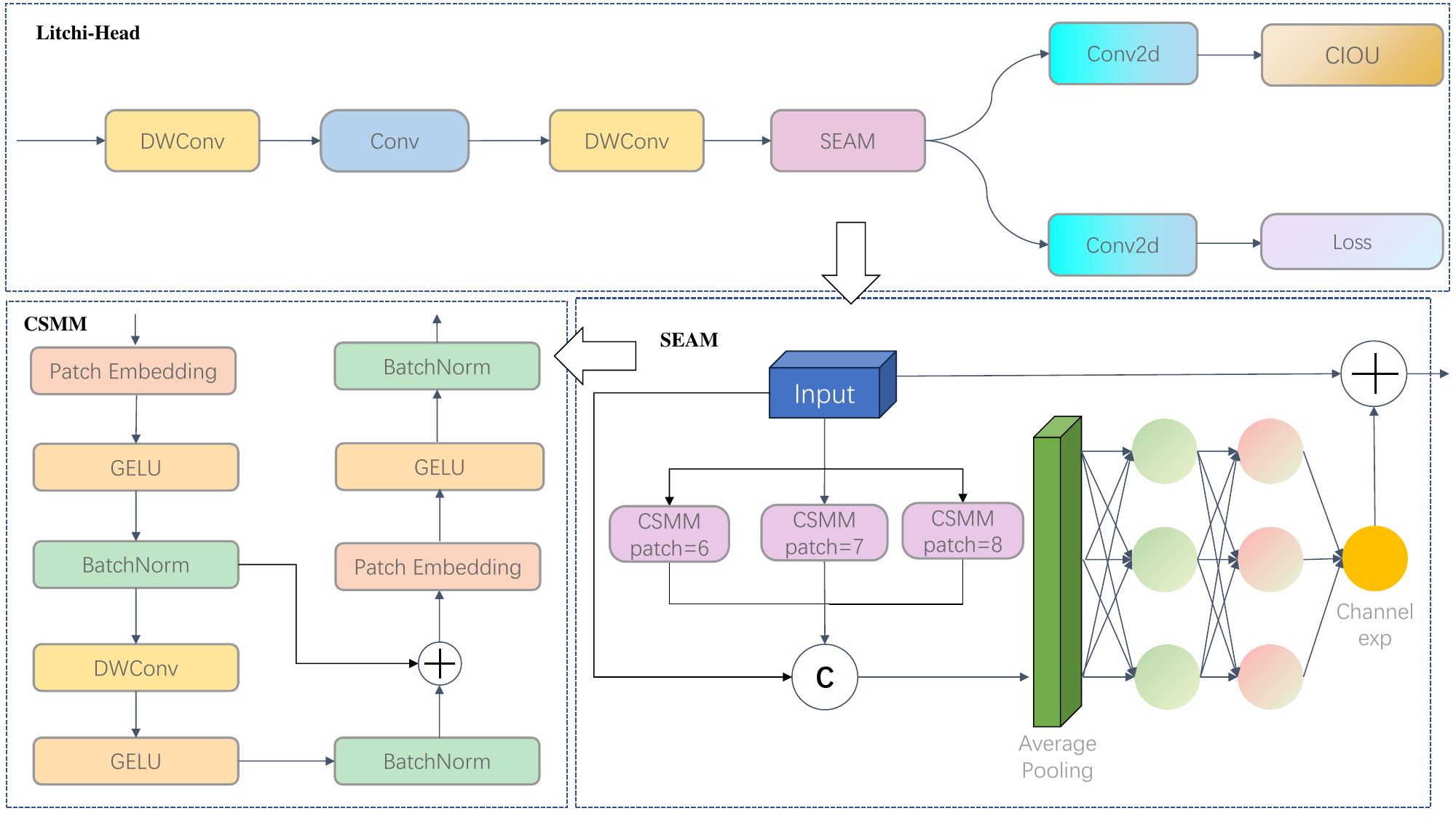}
    \caption{The details of Litchi-Head.}
    \label{fig:head}
\end{figure}

In complex scenes, occlusion frequently occurs, leading to alignment errors, local aliasing, and feature loss, which are major factors that compromise target detection accuracy. Litchi fruits, due to their cluster growth patterns, are particularly prone to mutual occlusion. Additionally, when UAVs capture images from an oblique angle at a certain altitude, litchi fruits are often obscured by branches and leaves, further complicating detection. To address these challenges, we propose the Litchi Occlusion Detection Head(Litchi-Head), an enhancement of the YOLOv11 detector that incorporates an occlusion attention mechanism. The details of Litchi-Head are illustrated in Figure~\ref{fig:head}.

Compared to the original YOLOv11 detector, the Litchi-Head adopts a parameter-sharing strategy by merging the two original branches into a single branch. This streamlined architecture allows input data to directly pass through two depthwise separable convolutions (DWConv)\cite{chollet2017xception} followed by a \(3 \times 3\) convolution. By avoiding redundant parameter transmission and storage, this design reduces the complexity of the model while maintaining computational efficiency.

To further enhance the model’s ability to handle occlusion, we introduce the Spatial-Enhanced Attention Module(SEAM)\cite{yu2024yolo}, which emphasizes litchi regions in the image while suppressing background noise. This module processes the input through three parallel branches, each employing a Channel and Space Hybrid Module (CSMM). Within each CSMM, the input is partitioned into patches of sizes \(6 \times 6\), \(7 \times 7\), and \(8 \times 8\) using the Patch Embedding method. These patches, each containing partial image information, are embedded into vector spaces for feature extraction. This multi-scale processing ensures the effective capture of diverse spatial features.

Each branch applies DWConv to learn spatial and channel correlations while minimizing the number of parameters. Although this method efficiently identifies the importance of different channels, it can overlook inter-channel relationships, leading to potential information loss. To address this limitation, the outputs from convolutions of varying depths are merged using a \(1 \times 1\) convolution, followed by a two-layer fully connected network. This design strengthens the connections between channels and compensates for information loss, particularly under occlusion scenarios. The relationship between occluded and non-occluded litchi fruits is further refined through this process, enabling the model to learn complex correlations effectively.

To improve tolerance for positional errors, the logits produced by the fully connected network are normalized using an exponential function, which maps the range from \([0, 1]\) to \([1, e]\). This normalization provides a monotonic mapping that enhances robustness against occlusion-induced inaccuracies. Finally, the refined features are combined with the original input through a residual connection, preserving the initial feature information while incorporating attention weights.

By emphasizing the litchi regions, strengthening inter-channel relationships, and enhancing positional robustness, the Litchi-Head effectively addresses the challenges posed by occlusion in UAV-based litchi detection tasks. The integration of these improvements enables the model to maintain high detection accuracy even in complex and cluttered environments.

\section{Experiments and discussion}\label{sec4}

\subsection{Experimental environment}
The experiments in this study were conducted on an Ubuntu 20.04 operating system, leveraging CUDA 12.3 and PyTorch 1.12.1 frameworks to handle deep learning tasks. Model training was performed on an NVIDIA GeForce RTX 3090 GPU with 24 GB of memory, ensuring efficient computation and high-performance processing. The hyperparameter configurations used in the experiments on the litchi-UAV dataset are detailed in Table~\ref{tab:hyper}.

\begin{table}
    \centering
    \begin{tabular}{>{\centering\arraybackslash}p{0.5\columnwidth} >{\centering\arraybackslash}p{0.4\columnwidth}}
         \toprule
         \bfseries Hyperparameters &  \bfseries Value \\
         \midrule
         Learning Rate & 0.01 \\
         Image Size & 1024x1024 \\
         Momentum & 0.937 \\
         Optimizer & SGD \\
         Batch Size & 16 \\
         Epoch & 300 \\
         Workers & 4 \\
         Weight Decay & 0.0005 \\
         \bottomrule
    \end{tabular}
    \caption{Hyperparameter configuration for experiments on the litchi-UAV dataset.}
    \label{tab:hyper}
\end{table}

\subsection{Evaluation index}
This study employs commonly used evaluation metrics for object detection models, including model parameters (Params), floating-point operations (GFLOPs), frames per second (FPS), precision (P), recall (R), F1-Score, and mean average precision (mAP). 

Params measure the storage space required by the model. A lower Params value indicates a lighter model, making it more suitable for deployment on mobile or embedded devices. GFLOPs quantify the computational resources and execution time needed for model operation, with lower values reflecting reduced resource consumption. FPS evaluates the model's processing speed in terms of frames per second, where a higher value signifies faster detection. For industrial real-time applications, an FPS greater than 30 is generally sufficient.

P, R, F1-Score, and mAP assess the detection performance of the model. Precision (P) measures the rate of false positives, indicating the proportion of correct predictions among all detections. Recall (R) measures the rate of missed detections, representing the proportion of actual targets correctly identified. F1-Score provides a balanced metric that combines precision and recall, offering a holistic measure of detection quality. Higher F1-Score values indicate better overall performance. Mean average precision (mAP) evaluates the algorithm's detection capability across all categories, providing a comprehensive performance metric. mAP is reported as mAP@50 and mAP@50:95, representing average precision at IoU thresholds of 50\% and 50\%-95\%, respectively. Higher mAP values correspond to better detection accuracy.

The mathematical definitions of P, R, F1-Score, and mAP are provided below, where TP represents true positives (correct detections), FP represents false positives (incorrect detections), FN represents false negatives (missed detections), and $q$ denotes the total number of classes.

\begin{equation}
P = \frac{TP}{TP + FP}
\end{equation}

\begin{equation}
R = \frac{TP}{TP + FN}
\end{equation}

\begin{equation}
F1\text{-Score} = \frac{2 \times P \times R}{P + R}
\end{equation}

\begin{equation}
mAP = \frac{\sum_{i=1}^{q} P(R_i) \, dR_i}{q}
\end{equation}

\subsection{Ablation experiment}

\begin{figure}
    \centering
    \includegraphics[width=0.95\linewidth]{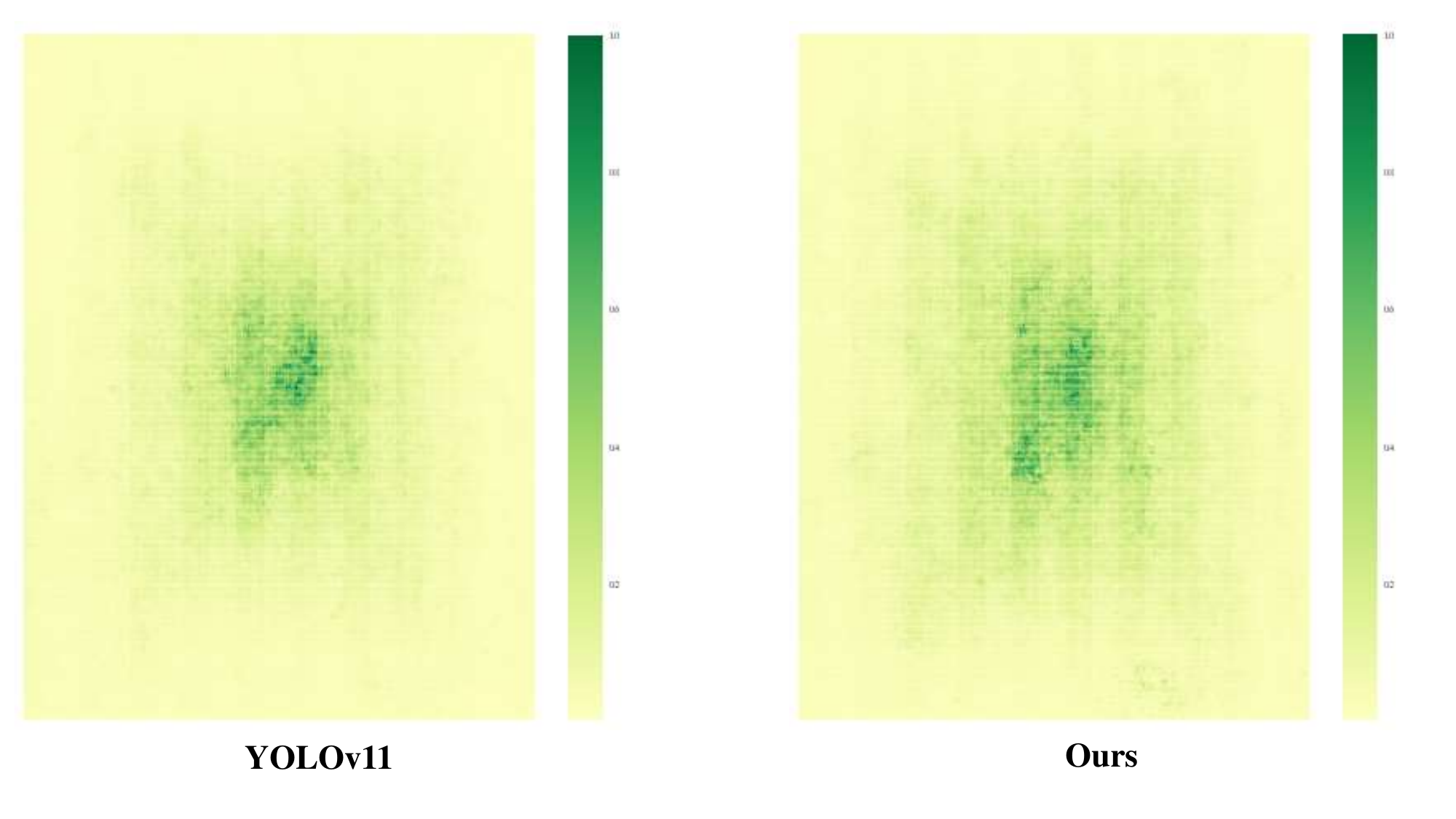}
    \caption{Receptive field visualization of the model}
    \label{fig:erf}
\end{figure}

The ablation experiments conducted in this study used the YOLOv11 model as the baseline, owing to its balance of high accuracy and low parameter requirements. As detailed in Chapter~\ref{sec3}, several enhancements were proposed to address challenges specific to litchi detection in UAV imagery. These include employing the C3-MSR module in the backbone to address the issue of small litchi targets, introducing a lightweight feature fusion method in the neck to enable deployment on mobile hardware, and incorporating the Litchi-Head module to mitigate the impact of occlusion on detection accuracy. 

The ablation experiments evaluated the performance of these modules individually and in combination, using Params, GFLOPs, P, R, F1-Score, and mAP@50 as evaluation metrics. The results are summarized in Table~\ref{tab:ablation}. 

The findings reveal that each proposed enhancement contributes to performance improvement when applied independently, with mAP@50 increasing by approximately 1\% for each module. Notably, the lightweight feature fusion method not only reduces Params and GFLOPs but also slightly improves mAP@50, demonstrating its ability to minimize computational overhead without compromising accuracy. Pairwise combinations of the modules yielded further improvements in mAP@50 and F1-Score compared to single-module implementations. When all three enhancements were integrated, the model achieved optimal performance across all metrics: mAP@50 improved by 2.4\%, Params were reduced to 6.35M (a 32.5\% reduction compared to the baseline), and the F1-Score reached 85.5\%. These results underscore the effectiveness of the proposed approach in enhancing model performance for litchi detection.

To further illustrate the improvements, the receptive field of the model was visualized. Figure~\ref{fig:erf} highlights the enhanced receptive field of the improved model compared to the baseline YOLOv11. The central green region in the visualization denotes the receptive field size. It is evident that the improved model has a significantly larger receptive field, resulting in stronger perceptual capability. This enhanced receptive field, combined with the proposed improvements, ensures the model's superior ability to learn and detect litchi targets effectively under challenging UAV scenarios.

\begin{table*}[htbp]
    \centering
    \begin{tabularx}{\textwidth}{>{\centering\arraybackslash}p{0.5cm} >{\centering\arraybackslash}p{0.5cm} >{\centering\arraybackslash}p{0.5cm} *{6}{>{\centering\arraybackslash}X}}
         \toprule
         \bfseries A & \bfseries B & \bfseries C & \bfseries Params (M) & \bfseries GFLOPs (G) & \bfseries $\bm{P(\%)}$ & \bfseries $\bm{R(\%)}$ & \bfseries F1-Score (\%) & \bfseries $\bm{mAP@50(\%)}$ \\
         \midrule
         $\times$ & $\times$ & $\times$ & 9.41 & 21.3 & 88.3 & 79.1 & 83.4 & 87.7\\
         $\bm{\checkmark}$ & $\times$ & $\times$ & 9.17 & 21.1 & 89.0 & 79.0 & 83.7 & 88.5\\
         $\times$ & $\bm{\checkmark}$ & $\times$ & 6.93 & 20.9 & 88.6 & 78.3 & 83.1 & 88.0\\
         $\times$ & $\times$ & $\bm{\checkmark}$ & 9.48 & 20.0 & 89.3 & 79.6 & 84.1 & 88.6\\
         $\bm{\checkmark}$ & $\bm{\checkmark}$ & $\times$ & 6.66 & 20.5 & 86.8 & 81.1 & 83.8 & 88.9\\
         $\bm{\checkmark}$ & $\times$ & $\bm{\checkmark}$ & 9.21 & 19.7 & 86.7 & 80.8 & 83.6 & 88.9\\
         $\times$ & $\bm{\checkmark}$ & $\bm{\checkmark}$ & 6.62 & 19.1 & 89.2 & 79.3 & 83.9 & 88.8\\
         $\bm{\checkmark}$ & $\bm{\checkmark}$ & $\bm{\checkmark}$ & $\bm{6.35}$ & $\bm{18.8}$ & $\bm{89.6}$ & $\bm{81.8}$ & $\bm{85.5}$ & $\bm{90.1}$\\
         \bottomrule
    \end{tabularx}
    \caption{Ablation experiments on parameters, GFLOPs, precision, recall, F1-Score, and mAP@50 for litchi-UAV detection. $A$: C3-MSR in the backbone; $B$: lightweight feature fusion in the neck; $C$: Litchi-Head replacing the YOLOv11 detection head.}
    \label{tab:ablation}
\end{table*}

\subsection{Model Performance Comparison Experiment}

\begin{figure}
    \centering
    \includegraphics[width=0.95\linewidth]{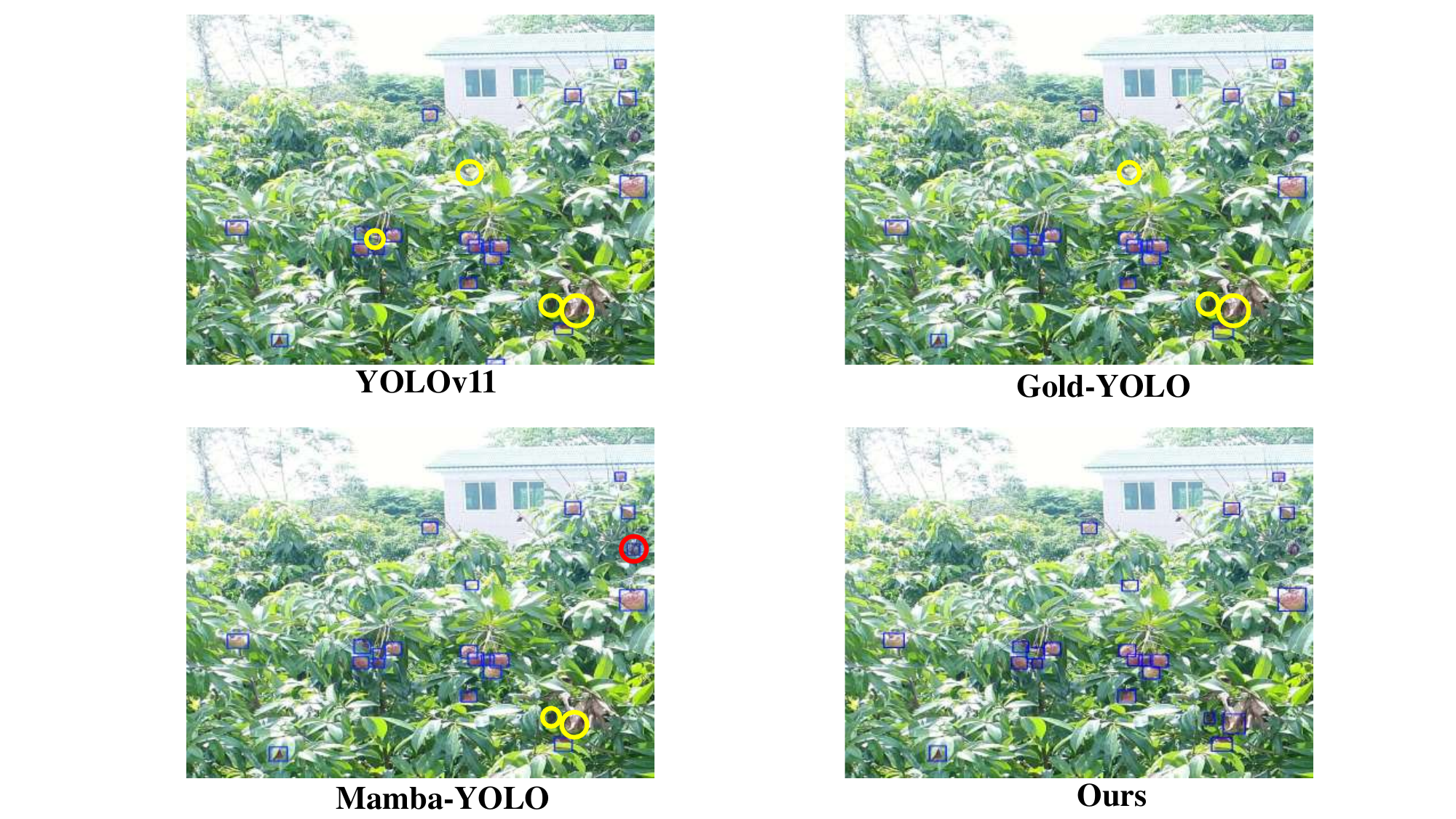}
    \caption{Visual comparison of model detection performance}
    \label{fig:visual}
\end{figure}

To evaluate the performance of our proposed model, we compared it against several widely used object detection models, all with parameter sizes not exceeding 20M. The selected models include YOLOv5 through YOLOv11, which are commonly utilized in various embedded scenarios. Additionally, to highlight the advantages of our model, we included two SOTA models specifically designed for object detection: Gold-YOLO and Mamba-YOLO\cite{wang2024mamba}. The evaluation metrics comprised Params, GFLOPs, FPS, P, R, F1-Score, and mAP@50. The results are summarized in Table~\ref{tab:compare}.

Our model achieved an mAP@50 of 90.1\% and an F1-Score of 85.5\% on the litchi-UAV dataset, surpassing all other models. Moreover, it demonstrated reduced parameters and GFLOPs, indicating that it can achieve superior detection performance with lower computational resources. While its FPS was slightly lower than some other models, it still exceeded the industrial real-time detection threshold of 30 FPS, ensuring practical applicability. These results highlight the effectiveness of our approach in addressing the unique challenges of litchi detection in UAV imagery.

To provide a more intuitive comparison, we visualized the detection results of YOLOv11, Gold-YOLO, and Mamba-YOLO alongside our model. The visualization is shown in Figure~\ref{fig:visual}. Due to the small size and high density of litchi fruits in the images, distinguishing them with the naked eye is challenging. To illustrate the comparative performance, yellow and red circles were used to mark missed and false detections, respectively. 

The figure reveals that Mamba-YOLO exhibited a lower missed detection rate compared to YOLOv11 and Gold-YOLO, successfully detecting litchi fruits obscured by leaves in the image's center. However, Mamba-YOLO encountered one instance of false detection. In contrast, our model not only effectively detected occluded litchi fruits, such as those in the lower-right corner of the image, but also achieved reduced rates of both missed and false detections, demonstrating its robustness and superior detection capabilities.

\begin{table*}[htbp]
    \centering
    \begin{tabularx}{\textwidth}{*{8}{>{\centering\arraybackslash}X}}
         \toprule
         \bfseries Model & \bfseries Params (M) & \bfseries GFLOPs (G) & \bfseries FPS & \bfseries $\bm{P(\%)}$ & \bfseries $\bm{R(\%)}$ & \bfseries F1-Score (\%) & \bfseries $\bm{mAP@50(\%)}$ \\
         \midrule
         YOLOv5s & 7.84 & 18.9 & $\bm{204.1}$ & 86.1 & $\bm{82.2}$ & 84.1 & 87.6\\
         YOLOv6s & 16.01 & 42.9 & 156.5 & $\bm{90.7}$ & 75.9 & 82.6 & 86.6\\
         YOLOv8s & 9.85 & 23.5 & 187.0 & 87.4 & 81.5 & 84.3 & 88.0\\
         YOLOv9s & 7.22 & 22.2 & 89.9 & 87.5 & 77.8 & 82.3 & 87.0\\
         YOLOv10s & 8.11 & 22.1 & 177.6 & 88.7 & 80.6 & 84.4 & 88.3\\
         YOLOv11s & 9.41 & 21.3 & 172.0 & 88.3 & 79.1 & 83.4 & 87.7\\
         GoldYOLO & 12.46 & 25.4 & 130.2 & 90.0 & 76.7 & 82.8 & 88.3\\
         MambaYOLO & 6.69 & 19.5 & 53.2 & 89.9 & 78.8 & 84.0 & 88.8\\
         \bfseries Ours & $\bm{6.35}$ & $\bm{18.8}$ & 57.2 & 89.6 & 81.8 & $\bm{85.5}$ & $\bm{90.1}$\\
         \bottomrule
    \end{tabularx}
    \caption{Performance comparison of various detection algorithms on the litchi-UAV dataset.}
    \label{tab:compare}
\end{table*}

\subsection{Comparative Experiment of Occlusion Detection Ability}

From the UAV's perspective, litchi fruits are frequently occluded by other fruits, branches, or leaves, making detection particularly challenging. In this study, occlusion scenarios were categorized into three types: fruit occlusion, non-occlusion, and branch or leaf occlusion. To evaluate the detection capability of our model under these different occlusion conditions, experiments were conducted to assess litchi fruit detection performance. The test results for five models are presented in Table~\ref{tab:occlusion}.

\begin{table*}
    \centering
    \begin{tabularx}{\textwidth}{*{7}{>{\centering\arraybackslash}X}}
         \toprule
         \multirow{2}{*}{\bfseries Model} & \multicolumn{2}{c}{\bfseries Fruit Occlusion} & \multicolumn{2}{c}{\bfseries Non-Occlusion} & \multicolumn{2}{c}{\bfseries Branch or Leaf Occlusion} \\
         & \bfseries Actual & \bfseries Undetected & \bfseries Actual & \bfseries Undetected & \bfseries Actual & \bfseries Undetected \\
         \midrule
         YOLOv10s & 185 & 34 & 922 & 169 & 278 & 55 \\
         YOLOv11s & 185 & 36 & 922 & 174 & 278 & 52 \\
         GoldYOLO & 185 & 28 & 922 & 141 & 278 & 45 \\
         MambaYOLO & 185 & 25 & 922 & 125 & 278 & 39 \\
         \bfseries Ours & 185 & $\bm{18}$ & 922 & $\bm{91}$ & 278 & $\bm{30}$ \\
         \bottomrule
    \end{tabularx}
    \caption{Comparison of occlusion detection ability under different conditions.}
    \label{tab:occlusion}
\end{table*}

Based on the results in Table~\ref{tab:occlusion}, our model YOLOv11-Litchi demonstrates the ability to detect at least 78\% of litchi fruits across all three occlusion scenarios. For fruit occlusion, the missed detection rates for YOLOv10s, YOLOv11s, Gold-YOLO, and Mamba-YOLO were 18.3\%, 19.4\%, 15.1\%, and 13.5\%, respectively. The relatively high rates of missed detections can be attributed to the similar color features and blurred contour boundaries of occluded fruits. In contrast, our model achieved a significantly lower missed detection rate of 9.7\% in the same scenario.

For branch or leaf occlusion, YOLOv10s, YOLOv11s, Gold-YOLO, and Mamba-YOLO exhibited missed detection rates of 19.7\%, 18.7\%, 16.1\%, and 14.0\%, respectively. However, our model achieved a missed detection rate of only 10.7\%, demonstrating a notable improvement over the other models. Under non-occlusion conditions, our model also outperformed the other methods, with a missed detection rate of just 9.9\%.

These findings highlight the robustness and superior performance of our model in detecting litchi fruits under various occlusion scenarios, effectively addressing challenges posed by overlapping fruits, branches, and leaves.

\subsection{Model Generalization Experiment}

To evaluate the generalization capability of our proposed model in crop image detection, we conducted experiments on two publicly available datasets: Laboro Tomato and Citrus. The hyperparameter configurations used in the experiments are detailed in Table~\ref{tab:hyper}. The experimental results are presented in Table~\ref{tab:gen1} and Table~\ref{tab:gen2}.

From the results, our model consistently outperformed other target detection models on both datasets, achieving higher scores across multiple evaluation metrics. Visualizations of the detection results on these datasets further illustrate the superior generalization ability of our model. The experiments demonstrate that our approach is highly effective in adapting to diverse crop image datasets, making it suitable for a wide range of agricultural applications.

\begin{table*}
    \centering
    \begin{tabularx}{\textwidth}{*{6}{>{\centering\arraybackslash}X}}
         \toprule
         \bfseries Model & \bfseries $\bm{P(\%)}$ & \bfseries $\bm{R(\%)}$ & \bfseries $\bm{F1-Score(\%)}$ & \bfseries $\bm{mAP@50(\%)}$ & \bfseries $\bm{mAP@50-95(\%)}$ \\
         \midrule
         YOLOv5s & 81.0 & 74.8 & 77.8 & 82.7 & 69.5 \\
         YOLOv6s & 77.1 & 77.0 & 77.1 & 82.7 & 69.5 \\
         YOLOv8s & 79.8 & 77.3 & 78.5 & 83.6 & 70.2 \\
         YOLOv9s & 81.0 & 76.9 & 78.9 & 84.6 & 71.2 \\
         YOLOv10s & 76.1 & 76.5 & 76.2 & 82.4 & 68.7 \\
         YOLOv11s & 81.5 & 77.6 & 79.5 & 83.9 & 70.5 \\
         GoldYOLO & 80.5 & 76.7 & 78.5 & 83.5 & 70.0 \\
         MambaYOLO & 77.2 & 74.4 & 75.7 & 81.6 & 66.7 \\
         \bfseries Ours & $\bm{81.8}$ & $\bm{80.7}$ & $\bm{81.2}$ & $\bm{84.9}$ & $\bm{71.4}$ \\
         \bottomrule
    \end{tabularx}
    \caption{Performance comparison of different models on the Laboro Tomato dataset.}
    \label{tab:gen1}
\end{table*}

\begin{table*}
    \centering
    \begin{tabularx}{\textwidth}{*{6}{>{\centering\arraybackslash}X}}
         \toprule
         \bfseries Model & \bfseries $\bm{P(\%)}$ & \bfseries $\bm{R(\%)}$ & \bfseries $\bm{F1-Score(\%)}$ & \bfseries $\bm{mAP@50(\%)}$ & \bfseries $\bm{mAP@50-95(\%)}$ \\
         \midrule
         YOLOv5s & 84.7 & 89.2 & 86.8 & 93.7 & 77.2 \\
         YOLOv6s & 84.9 & 89.2 & 86.9 & 93.2 & 76.8 \\
         YOLOv8s & 84.2 & 88.8 & 86.4 & 93.2 & 76.9 \\
         YOLOv9s & 86.3 & 88.1 & 87.2 & 93.4 & 77.4 \\
         YOLOv10s & $\bm{87.1}$ & 86.1 & 86.6 & 93.5 & 76.8 \\
         YOLOv11s & 85.6 & 88.7 & 87.1 & 93.3 & 77.6 \\
         GoldYOLO & 85.1 & 89.2 & 87.1 & 94.0 & 76.9 \\
         MambaYOLO & 85.5 & 89.4 & 87.4 & 94.2 & 77.1 \\
         \bfseries Ours & 85.6 & $\bm{89.6}$ & $\bm{87.5}$ & $\bm{94.5}$ & $\bm{77.6}$ \\
         \bottomrule
    \end{tabularx}
    \caption{Performance comparison of different models on the Citrus dataset.}
    \label{tab:gen2}
\end{table*}

\section{Conclusion}\label{sec5}

To address the challenges of detecting litchi in UAV imagery, the difficulties associated with deploying models with large parameters, and the frequent target occlusion in UAV-based litchi detection tasks, this paper proposes an improved detection model YOLOv11-Litchi. The model integrates several key strategies, including a multi-scale residual module, a lightweight feature fusion method, and a litchi occlusion detection head.

Firstly, the multi-scale residual module is introduced to enhance the efficiency of multi-scale feature fusion, effectively capturing contextual information across different scales. Secondly, to facilitate deployment on UAV platforms, a lightweight feature fusion method is designed to significantly reduce model parameters and computational complexity while maintaining high detection accuracy. Finally, a litchi occlusion detection head is proposed to focus on litchi regions in the image, suppress background interference, and mitigate the adverse effects of occlusion on detection performance.

Experimental results validate the effectiveness of the proposed model. The model achieves a parameter size of 6.35 MB, which is 32.5\% smaller than the YOLOv11 benchmark network, while improving the mAP by 2.5\%, reaching 90.1\%. The F1-Score is also increased by 1.4\%, reaching 85.5\%. Additionally, the model achieves a frame rate of 57.2 FPS, meeting the requirements for real-time performance and achieving an optimal balance between accuracy and speed. Generalization experiments further demonstrate the robustness and adaptability of the model to other crop detection tasks, highlighting its potential for broader applications in precision agriculture.

\section{Acknowledgments}
This research was supported by Guangzhou Science and Technology Plan Project (Grant No. 2024E04J1242, 2023B01J0046) and Natural Science Foundation of China (Grant No. 61863011, 32071912).

\printcredits

\bibliographystyle{cas-model2-names}

\bibliography{cas-refs}

\begin{thebibliography}{45}
\expandafter\ifx\csname natexlab\endcsname\relax\def\natexlab#1{#1}\fi
\providecommand{\url}[1]{\texttt{#1}}
\providecommand{\href}[2]{#2}
\providecommand{\path}[1]{#1}
\providecommand{\DOIprefix}{doi:}
\providecommand{\ArXivprefix}{arXiv:}
\providecommand{\URLprefix}{URL: }
\providecommand{\Pubmedprefix}{pmid:}
\providecommand{\doi}[1]{\href{http://dx.doi.org/#1}{\path{#1}}}
\providecommand{\Pubmed}[1]{\href{pmid:#1}{\path{#1}}}
\providecommand{\bibinfo}[2]{#2}
\ifx\xfnm\relax \def\xfnm[#1]{\unskip,\space#1}\fi
\bibitem[{Alwateer et~al.(2019)Alwateer, Loke and Fernando}]{alwateer2019enabling}
\bibinfo{author}{Alwateer, M.}, \bibinfo{author}{Loke, S.W.}, \bibinfo{author}{Fernando, N.}, \bibinfo{year}{2019}.
\newblock \bibinfo{title}{Enabling drone services: Drone crowdsourcing and drone scripting}.
\newblock \bibinfo{journal}{IEEE access} \bibinfo{volume}{7}, \bibinfo{pages}{110035--110049}.
\bibitem[{Ashish(2017)}]{ashish2017attention}
\bibinfo{author}{Ashish, V.}, \bibinfo{year}{2017}.
\newblock \bibinfo{title}{Attention is all you need}.
\newblock \bibinfo{journal}{Advances in neural information processing systems} \bibinfo{volume}{30}, \bibinfo{pages}{I}.
\bibitem[{Auernhammer(2001)}]{auernhammer2001precision}
\bibinfo{author}{Auernhammer, H.}, \bibinfo{year}{2001}.
\newblock \bibinfo{title}{Precision farming—the environmental challenge}.
\newblock \bibinfo{journal}{Computers and electronics in agriculture} \bibinfo{volume}{30}, \bibinfo{pages}{31--43}.
\bibitem[{Azizi et~al.(2024)Azizi, Zhang, Rui, Li, Igathinathane, Flores, Mathew, Pourreza, Han and Zhang}]{azizi2024comprehensive}
\bibinfo{author}{Azizi, A.}, \bibinfo{author}{Zhang, Z.}, \bibinfo{author}{Rui, Z.}, \bibinfo{author}{Li, Y.}, \bibinfo{author}{Igathinathane, C.}, \bibinfo{author}{Flores, P.}, \bibinfo{author}{Mathew, J.}, \bibinfo{author}{Pourreza, A.}, \bibinfo{author}{Han, X.}, \bibinfo{author}{Zhang, M.}, \bibinfo{year}{2024}.
\newblock \bibinfo{title}{Comprehensive wheat lodging detection after initial lodging using uav rgb images}.
\newblock \bibinfo{journal}{Expert Systems with Applications} \bibinfo{volume}{238}, \bibinfo{pages}{121788}.
\bibitem[{Cai et~al.(2016)Cai, Fan, Feris and Vasconcelos}]{cai2016unified}
\bibinfo{author}{Cai, Z.}, \bibinfo{author}{Fan, Q.}, \bibinfo{author}{Feris, R.S.}, \bibinfo{author}{Vasconcelos, N.}, \bibinfo{year}{2016}.
\newblock \bibinfo{title}{A unified multi-scale deep convolutional neural network for fast object detection}, in: \bibinfo{booktitle}{Computer Vision--ECCV 2016: 14th European Conference, Amsterdam, The Netherlands, October 11--14, 2016, Proceedings, Part IV 14}, \bibinfo{organization}{Springer}. pp. \bibinfo{pages}{354--370}.
\bibitem[{Carion et~al.(2020)Carion, Massa, Synnaeve, Usunier, Kirillov and Zagoruyko}]{carion2020end}
\bibinfo{author}{Carion, N.}, \bibinfo{author}{Massa, F.}, \bibinfo{author}{Synnaeve, G.}, \bibinfo{author}{Usunier, N.}, \bibinfo{author}{Kirillov, A.}, \bibinfo{author}{Zagoruyko, S.}, \bibinfo{year}{2020}.
\newblock \bibinfo{title}{End-to-end object detection with transformers}, in: \bibinfo{booktitle}{European conference on computer vision}, \bibinfo{organization}{Springer}. pp. \bibinfo{pages}{213--229}.
\bibitem[{Chen et~al.(2023)Chen, Kao, He, Zhuo, Wen, Lee and Chan}]{chen2023run}
\bibinfo{author}{Chen, J.}, \bibinfo{author}{Kao, S.h.}, \bibinfo{author}{He, H.}, \bibinfo{author}{Zhuo, W.}, \bibinfo{author}{Wen, S.}, \bibinfo{author}{Lee, C.H.}, \bibinfo{author}{Chan, S.H.G.}, \bibinfo{year}{2023}.
\newblock \bibinfo{title}{Run, don't walk: chasing higher flops for faster neural networks}, in: \bibinfo{booktitle}{Proceedings of the IEEE/CVF conference on computer vision and pattern recognition}, pp. \bibinfo{pages}{12021--12031}.
\bibitem[{Chen et~al.(2021)Chen, Chang, Hsieh and Chen}]{chen2021parallel}
\bibinfo{author}{Chen, P.Y.}, \bibinfo{author}{Chang, M.C.}, \bibinfo{author}{Hsieh, J.W.}, \bibinfo{author}{Chen, Y.S.}, \bibinfo{year}{2021}.
\newblock \bibinfo{title}{Parallel residual bi-fusion feature pyramid network for accurate single-shot object detection}.
\newblock \bibinfo{journal}{IEEE transactions on Image Processing} \bibinfo{volume}{30}, \bibinfo{pages}{9099--9111}.
\bibitem[{Chollet(2017)}]{chollet2017xception}
\bibinfo{author}{Chollet, F.}, \bibinfo{year}{2017}.
\newblock \bibinfo{title}{Xception: Deep learning with depthwise separable convolutions}, in: \bibinfo{booktitle}{Proceedings of the IEEE conference on computer vision and pattern recognition}, pp. \bibinfo{pages}{1251--1258}.
\bibitem[{Cui et~al.(2024)Cui, Zhang, Zhang, Han, Ai, Dong and Liu}]{cui2024weed}
\bibinfo{author}{Cui, J.}, \bibinfo{author}{Zhang, X.}, \bibinfo{author}{Zhang, J.}, \bibinfo{author}{Han, Y.}, \bibinfo{author}{Ai, H.}, \bibinfo{author}{Dong, C.}, \bibinfo{author}{Liu, H.}, \bibinfo{year}{2024}.
\newblock \bibinfo{title}{Weed identification in soybean seedling stage based on uav images and faster r-cnn}.
\newblock \bibinfo{journal}{Computers and Electronics in Agriculture} \bibinfo{volume}{227}, \bibinfo{pages}{109533}.
\bibitem[{Ding et~al.(2022a)Ding, Chen, Zhang, Huang, Han and Ding}]{ding2022re}
\bibinfo{author}{Ding, X.}, \bibinfo{author}{Chen, H.}, \bibinfo{author}{Zhang, X.}, \bibinfo{author}{Huang, K.}, \bibinfo{author}{Han, J.}, \bibinfo{author}{Ding, G.}, \bibinfo{year}{2022}a.
\newblock \bibinfo{title}{Re-parameterizing your optimizers rather than architectures}.
\newblock \bibinfo{journal}{arXiv preprint arXiv:2205.15242} .
\bibitem[{Ding et~al.(2021a)Ding, Zhang, Han and Ding}]{ding2021diverse}
\bibinfo{author}{Ding, X.}, \bibinfo{author}{Zhang, X.}, \bibinfo{author}{Han, J.}, \bibinfo{author}{Ding, G.}, \bibinfo{year}{2021}a.
\newblock \bibinfo{title}{Diverse branch block: Building a convolution as an inception-like unit}, in: \bibinfo{booktitle}{Proceedings of the IEEE/CVF conference on computer vision and pattern recognition}, pp. \bibinfo{pages}{10886--10895}.
\bibitem[{Ding et~al.(2022b)Ding, Zhang, Han and Ding}]{ding2022scaling}
\bibinfo{author}{Ding, X.}, \bibinfo{author}{Zhang, X.}, \bibinfo{author}{Han, J.}, \bibinfo{author}{Ding, G.}, \bibinfo{year}{2022}b.
\newblock \bibinfo{title}{Scaling up your kernels to 31x31: Revisiting large kernel design in cnns}, in: \bibinfo{booktitle}{Proceedings of the IEEE/CVF conference on computer vision and pattern recognition}, pp. \bibinfo{pages}{11963--11975}.
\bibitem[{Ding et~al.(2021b)Ding, Zhang, Ma, Han, Ding and Sun}]{ding2021repvgg}
\bibinfo{author}{Ding, X.}, \bibinfo{author}{Zhang, X.}, \bibinfo{author}{Ma, N.}, \bibinfo{author}{Han, J.}, \bibinfo{author}{Ding, G.}, \bibinfo{author}{Sun, J.}, \bibinfo{year}{2021}b.
\newblock \bibinfo{title}{Repvgg: Making vgg-style convnets great again}, in: \bibinfo{booktitle}{Proceedings of the IEEE/CVF conference on computer vision and pattern recognition}, pp. \bibinfo{pages}{13733--13742}.
\bibitem[{Ding et~al.(2024)Ding, Zhang, Ge, Zhao, Song, Yue and Shan}]{ding2024unireplknet}
\bibinfo{author}{Ding, X.}, \bibinfo{author}{Zhang, Y.}, \bibinfo{author}{Ge, Y.}, \bibinfo{author}{Zhao, S.}, \bibinfo{author}{Song, L.}, \bibinfo{author}{Yue, X.}, \bibinfo{author}{Shan, Y.}, \bibinfo{year}{2024}.
\newblock \bibinfo{title}{Unireplknet: A universal perception large-kernel convnet for audio video point cloud time-series and image recognition}, in: \bibinfo{booktitle}{Proceedings of the IEEE/CVF Conference on Computer Vision and Pattern Recognition}, pp. \bibinfo{pages}{5513--5524}.
\bibitem[{Du et~al.(2023)Du, Cheng, Ma, Lu, Wang, Meng, Jiang and Hong}]{du2023dsw}
\bibinfo{author}{Du, X.}, \bibinfo{author}{Cheng, H.}, \bibinfo{author}{Ma, Z.}, \bibinfo{author}{Lu, W.}, \bibinfo{author}{Wang, M.}, \bibinfo{author}{Meng, Z.}, \bibinfo{author}{Jiang, C.}, \bibinfo{author}{Hong, F.}, \bibinfo{year}{2023}.
\newblock \bibinfo{title}{Dsw-yolo: A detection method for ground-planted strawberry fruits under different occlusion levels}.
\newblock \bibinfo{journal}{Computers and Electronics in Agriculture} \bibinfo{volume}{214}, \bibinfo{pages}{108304}.
\bibitem[{Gao et~al.(2024)Gao, Liao, Nuyttens, Lootens, Xue, Alexandersson and Pieters}]{gao2024cross}
\bibinfo{author}{Gao, J.}, \bibinfo{author}{Liao, W.}, \bibinfo{author}{Nuyttens, D.}, \bibinfo{author}{Lootens, P.}, \bibinfo{author}{Xue, W.}, \bibinfo{author}{Alexandersson, E.}, \bibinfo{author}{Pieters, J.}, \bibinfo{year}{2024}.
\newblock \bibinfo{title}{Cross-domain transfer learning for weed segmentation and mapping in precision farming using ground and uav images}.
\newblock \bibinfo{journal}{Expert Systems with applications} \bibinfo{volume}{246}, \bibinfo{pages}{122980}.
\bibitem[{Gen{\'e}-Mola et~al.(2020)Gen{\'e}-Mola, Sanz-Cortiella, Rosell-Polo, Morros, Ruiz-Hidalgo, Vilaplana and Gregorio}]{gene2020fruit}
\bibinfo{author}{Gen{\'e}-Mola, J.}, \bibinfo{author}{Sanz-Cortiella, R.}, \bibinfo{author}{Rosell-Polo, J.R.}, \bibinfo{author}{Morros, J.R.}, \bibinfo{author}{Ruiz-Hidalgo, J.}, \bibinfo{author}{Vilaplana, V.}, \bibinfo{author}{Gregorio, E.}, \bibinfo{year}{2020}.
\newblock \bibinfo{title}{Fruit detection and 3d location using instance segmentation neural networks and structure-from-motion photogrammetry}.
\newblock \bibinfo{journal}{Computers and Electronics in Agriculture} \bibinfo{volume}{169}, \bibinfo{pages}{105165}.
\bibitem[{He et~al.(2017)He, Gkioxari, Doll{\'a}r and Girshick}]{he2017mask}
\bibinfo{author}{He, K.}, \bibinfo{author}{Gkioxari, G.}, \bibinfo{author}{Doll{\'a}r, P.}, \bibinfo{author}{Girshick, R.}, \bibinfo{year}{2017}.
\newblock \bibinfo{title}{Mask r-cnn}, in: \bibinfo{booktitle}{Proceedings of the IEEE international conference on computer vision}, pp. \bibinfo{pages}{2961--2969}.
\bibitem[{He et~al.(2016)He, Zhang, Ren and Sun}]{he2016deep}
\bibinfo{author}{He, K.}, \bibinfo{author}{Zhang, X.}, \bibinfo{author}{Ren, S.}, \bibinfo{author}{Sun, J.}, \bibinfo{year}{2016}.
\newblock \bibinfo{title}{Deep residual learning for image recognition}, in: \bibinfo{booktitle}{Proceedings of the IEEE conference on computer vision and pattern recognition}, pp. \bibinfo{pages}{770--778}.
\bibitem[{Hou et~al.(2022)Hou, Zhang, Tang, Zhuang, Tan, Huang, Chen, Wei, He and Luo}]{hou2022detection}
\bibinfo{author}{Hou, C.}, \bibinfo{author}{Zhang, X.}, \bibinfo{author}{Tang, Y.}, \bibinfo{author}{Zhuang, J.}, \bibinfo{author}{Tan, Z.}, \bibinfo{author}{Huang, H.}, \bibinfo{author}{Chen, W.}, \bibinfo{author}{Wei, S.}, \bibinfo{author}{He, Y.}, \bibinfo{author}{Luo, S.}, \bibinfo{year}{2022}.
\newblock \bibinfo{title}{Detection and localization of citrus fruit based on improved you only look once v5s and binocular vision in the orchard}.
\newblock \bibinfo{journal}{Frontiers in Plant Science} \bibinfo{volume}{13}, \bibinfo{pages}{972445}.
\bibitem[{Joshi et~al.(2024)Joshi, Sandhu, Dhillon, Chen and Bohara}]{joshi2024detection}
\bibinfo{author}{Joshi, P.}, \bibinfo{author}{Sandhu, K.S.}, \bibinfo{author}{Dhillon, G.S.}, \bibinfo{author}{Chen, J.}, \bibinfo{author}{Bohara, K.}, \bibinfo{year}{2024}.
\newblock \bibinfo{title}{Detection and monitoring wheat diseases using unmanned aerial vehicles (uavs)}.
\newblock \bibinfo{journal}{Computers and Electronics in Agriculture} \bibinfo{volume}{224}, \bibinfo{pages}{109158}.
\bibitem[{Kuang et~al.(2023)Kuang, Wang, Cheng, Li, Li, Zhang, Shen, Li and Xu}]{kuang2023residue}
\bibinfo{author}{Kuang, L.}, \bibinfo{author}{Wang, Z.}, \bibinfo{author}{Cheng, Y.}, \bibinfo{author}{Li, Y.}, \bibinfo{author}{Li, H.}, \bibinfo{author}{Zhang, J.}, \bibinfo{author}{Shen, Y.}, \bibinfo{author}{Li, J.}, \bibinfo{author}{Xu, G.}, \bibinfo{year}{2023}.
\newblock \bibinfo{title}{Residue levels and risk assessment of pesticides in litchi and longan of china}.
\newblock \bibinfo{journal}{Journal of Food Composition and Analysis} \bibinfo{volume}{115}, \bibinfo{pages}{104921}.
\bibitem[{LaboroAI(2024)}]{laboro_tomato}
\bibinfo{author}{LaboroAI}, \bibinfo{year}{2024}.
\newblock \bibinfo{title}{Laboro tomato dataset}.
\newblock \bibinfo{howpublished}{\url{https://github.com/laboroai/LaboroTomato}}.
\newblock \bibinfo{note}{Accessed: 2024-11-16}.
\bibitem[{Lee et~al.(2023)Lee, Yang, Tseng, Hsu, Sung and Chen}]{lee2023single}
\bibinfo{author}{Lee, C.J.}, \bibinfo{author}{Yang, M.D.}, \bibinfo{author}{Tseng, H.H.}, \bibinfo{author}{Hsu, Y.C.}, \bibinfo{author}{Sung, Y.}, \bibinfo{author}{Chen, W.L.}, \bibinfo{year}{2023}.
\newblock \bibinfo{title}{Single-plant broccoli growth monitoring using deep learning with uav imagery}.
\newblock \bibinfo{journal}{Computers and Electronics in Agriculture} \bibinfo{volume}{207}, \bibinfo{pages}{107739}.
\bibitem[{Li et~al.(2021)Li, Sun, Elkhouchlaa, Jia, Yao, Lin, Li and Lu}]{li2021fast}
\bibinfo{author}{Li, D.}, \bibinfo{author}{Sun, X.}, \bibinfo{author}{Elkhouchlaa, H.}, \bibinfo{author}{Jia, Y.}, \bibinfo{author}{Yao, Z.}, \bibinfo{author}{Lin, P.}, \bibinfo{author}{Li, J.}, \bibinfo{author}{Lu, H.}, \bibinfo{year}{2021}.
\newblock \bibinfo{title}{Fast detection and location of longan fruits using uav images}.
\newblock \bibinfo{journal}{Computers and Electronics in Agriculture} \bibinfo{volume}{190}, \bibinfo{pages}{106465}.
\bibitem[{Li et~al.(2024)Li, Shah, Xiong, Zhang and Wu}]{li2024unmanned}
\bibinfo{author}{Li, Z.}, \bibinfo{author}{Shah, F.}, \bibinfo{author}{Xiong, L.}, \bibinfo{author}{Zhang, J.}, \bibinfo{author}{Wu, W.}, \bibinfo{year}{2024}.
\newblock \bibinfo{title}{Unmanned aerial vehicles (uavs)-based crop lodging susceptibility and seed yield assessment during different growth stages of rapeseed (brassica napus)}.
\newblock \bibinfo{journal}{Computers and Electronics in Agriculture} \bibinfo{volume}{221}, \bibinfo{pages}{108980}.
\bibitem[{Liang et~al.(2024)Liang, Li, Wu, Zhao, Liu, Liu, Liu, Fan, Pan, Shen et~al.}]{liang2024rotated}
\bibinfo{author}{Liang, Y.}, \bibinfo{author}{Li, H.}, \bibinfo{author}{Wu, H.}, \bibinfo{author}{Zhao, Y.}, \bibinfo{author}{Liu, Z.}, \bibinfo{author}{Liu, D.}, \bibinfo{author}{Liu, Z.}, \bibinfo{author}{Fan, G.}, \bibinfo{author}{Pan, Z.}, \bibinfo{author}{Shen, Z.}, et~al., \bibinfo{year}{2024}.
\newblock \bibinfo{title}{A rotated rice spike detection model and a crop yield estimation application based on uav images}.
\newblock \bibinfo{journal}{Computers and Electronics in Agriculture} \bibinfo{volume}{224}, \bibinfo{pages}{109188}.
\bibitem[{Lin et~al.(2017)Lin, Doll{\'a}r, Girshick, He, Hariharan and Belongie}]{lin2017feature}
\bibinfo{author}{Lin, T.Y.}, \bibinfo{author}{Doll{\'a}r, P.}, \bibinfo{author}{Girshick, R.}, \bibinfo{author}{He, K.}, \bibinfo{author}{Hariharan, B.}, \bibinfo{author}{Belongie, S.}, \bibinfo{year}{2017}.
\newblock \bibinfo{title}{Feature pyramid networks for object detection}, in: \bibinfo{booktitle}{Proceedings of the IEEE conference on computer vision and pattern recognition}, pp. \bibinfo{pages}{2117--2125}.
\bibitem[{Liu et~al.(2018)Liu, Qi, Qin, Shi and Jia}]{liu2018path}
\bibinfo{author}{Liu, S.}, \bibinfo{author}{Qi, L.}, \bibinfo{author}{Qin, H.}, \bibinfo{author}{Shi, J.}, \bibinfo{author}{Jia, J.}, \bibinfo{year}{2018}.
\newblock \bibinfo{title}{Path aggregation network for instance segmentation}, in: \bibinfo{booktitle}{Proceedings of the IEEE conference on computer vision and pattern recognition}, pp. \bibinfo{pages}{8759--8768}.
\bibitem[{Liu et~al.(2016)Liu, Anguelov, Erhan, Szegedy, Reed, Fu and Berg}]{liu2016ssd}
\bibinfo{author}{Liu, W.}, \bibinfo{author}{Anguelov, D.}, \bibinfo{author}{Erhan, D.}, \bibinfo{author}{Szegedy, C.}, \bibinfo{author}{Reed, S.}, \bibinfo{author}{Fu, C.Y.}, \bibinfo{author}{Berg, A.C.}, \bibinfo{year}{2016}.
\newblock \bibinfo{title}{Ssd: Single shot multibox detector}, in: \bibinfo{booktitle}{Computer Vision--ECCV 2016: 14th European Conference, Amsterdam, The Netherlands, October 11--14, 2016, Proceedings, Part I 14}, \bibinfo{organization}{Springer}. pp. \bibinfo{pages}{21--37}.
\bibitem[{Najibi et~al.(2017)Najibi, Samangouei, Chellappa and Davis}]{najibi2017ssh}
\bibinfo{author}{Najibi, M.}, \bibinfo{author}{Samangouei, P.}, \bibinfo{author}{Chellappa, R.}, \bibinfo{author}{Davis, L.S.}, \bibinfo{year}{2017}.
\newblock \bibinfo{title}{Ssh: Single stage headless face detector}, in: \bibinfo{booktitle}{Proceedings of the IEEE international conference on computer vision}, pp. \bibinfo{pages}{4875--4884}.
\bibitem[{Ouyang et~al.(2023)Ouyang, He, Zhang, Luo, Guo, Zhan and Huang}]{ouyang2023efficient}
\bibinfo{author}{Ouyang, D.}, \bibinfo{author}{He, S.}, \bibinfo{author}{Zhang, G.}, \bibinfo{author}{Luo, M.}, \bibinfo{author}{Guo, H.}, \bibinfo{author}{Zhan, J.}, \bibinfo{author}{Huang, Z.}, \bibinfo{year}{2023}.
\newblock \bibinfo{title}{Efficient multi-scale attention module with cross-spatial learning}, in: \bibinfo{booktitle}{ICASSP 2023-2023 IEEE International Conference on Acoustics, Speech and Signal Processing (ICASSP)}, \bibinfo{organization}{IEEE}. pp. \bibinfo{pages}{1--5}.
\bibitem[{Qi et~al.(2022)Qi, Dong, Lan and Zhu}]{qi2022method}
\bibinfo{author}{Qi, X.}, \bibinfo{author}{Dong, J.}, \bibinfo{author}{Lan, Y.}, \bibinfo{author}{Zhu, H.}, \bibinfo{year}{2022}.
\newblock \bibinfo{title}{Method for identifying litchi picking position based on yolov5 and pspnet}.
\newblock \bibinfo{journal}{Remote Sensing} \bibinfo{volume}{14}, \bibinfo{pages}{2004}.
\bibitem[{Sun et~al.(2024)Sun, Zhang, Gao, Zhang, Li and Miao}]{sun2024efficient}
\bibinfo{author}{Sun, T.}, \bibinfo{author}{Zhang, W.}, \bibinfo{author}{Gao, X.}, \bibinfo{author}{Zhang, W.}, \bibinfo{author}{Li, N.}, \bibinfo{author}{Miao, Z.}, \bibinfo{year}{2024}.
\newblock \bibinfo{title}{Efficient occlusion avoidance based on active deep sensing for harvesting robots}.
\newblock \bibinfo{journal}{Computers and Electronics in Agriculture} \bibinfo{volume}{225}, \bibinfo{pages}{109360}.
\bibitem[{Tan et~al.(2020)Tan, Pang and Le}]{tan2020efficientdet}
\bibinfo{author}{Tan, M.}, \bibinfo{author}{Pang, R.}, \bibinfo{author}{Le, Q.V.}, \bibinfo{year}{2020}.
\newblock \bibinfo{title}{Efficientdet: Scalable and efficient object detection}, in: \bibinfo{booktitle}{Proceedings of the IEEE/CVF conference on computer vision and pattern recognition}, pp. \bibinfo{pages}{10781--10790}.
\bibitem[{Tetila et~al.(2020)Tetila, Machado, Astolfi, de~Souza~Belete, Amorim, Roel and Pistori}]{tetila2020detection}
\bibinfo{author}{Tetila, E.C.}, \bibinfo{author}{Machado, B.B.}, \bibinfo{author}{Astolfi, G.}, \bibinfo{author}{de~Souza~Belete, N.A.}, \bibinfo{author}{Amorim, W.P.}, \bibinfo{author}{Roel, A.R.}, \bibinfo{author}{Pistori, H.}, \bibinfo{year}{2020}.
\newblock \bibinfo{title}{Detection and classification of soybean pests using deep learning with uav images}.
\newblock \bibinfo{journal}{Computers and Electronics in Agriculture} \bibinfo{volume}{179}, \bibinfo{pages}{105836}.
\bibitem[{Wang et~al.(2024a)Wang, He, Nie, Guo, Liu, Wang and Han}]{wang2024gold}
\bibinfo{author}{Wang, C.}, \bibinfo{author}{He, W.}, \bibinfo{author}{Nie, Y.}, \bibinfo{author}{Guo, J.}, \bibinfo{author}{Liu, C.}, \bibinfo{author}{Wang, Y.}, \bibinfo{author}{Han, K.}, \bibinfo{year}{2024}a.
\newblock \bibinfo{title}{Gold-yolo: Efficient object detector via gather-and-distribute mechanism}.
\newblock \bibinfo{journal}{Advances in Neural Information Processing Systems} \bibinfo{volume}{36}.
\bibitem[{Wang et~al.(2017)Wang, Yuan and Yu}]{wang2017face}
\bibinfo{author}{Wang, J.}, \bibinfo{author}{Yuan, Y.}, \bibinfo{author}{Yu, G.}, \bibinfo{year}{2017}.
\newblock \bibinfo{title}{Face attention network: An effective face detector for the occluded faces}.
\newblock \bibinfo{journal}{arXiv preprint arXiv:1711.07246} .
\bibitem[{Wang et~al.(2021)Wang, Liu and Liu}]{wang2021diseases}
\bibinfo{author}{Wang, X.}, \bibinfo{author}{Liu, J.}, \bibinfo{author}{Liu, G.}, \bibinfo{year}{2021}.
\newblock \bibinfo{title}{Diseases detection of occlusion and overlapping tomato leaves based on deep learning}.
\newblock \bibinfo{journal}{Frontiers in plant science} \bibinfo{volume}{12}, \bibinfo{pages}{792244}.
\bibitem[{Wang et~al.(2024b)Wang, Li, Xu and Zhu}]{wang2024mamba}
\bibinfo{author}{Wang, Z.}, \bibinfo{author}{Li, C.}, \bibinfo{author}{Xu, H.}, \bibinfo{author}{Zhu, X.}, \bibinfo{year}{2024}b.
\newblock \bibinfo{title}{Mamba yolo: Ssms-based yolo for object detection}.
\newblock \bibinfo{journal}{arXiv preprint arXiv:2406.05835} .
\bibitem[{Wei et~al.(2022)Wei, Liu, Xu, Dai, Dai and Xu}]{wei2022dwrseg}
\bibinfo{author}{Wei, H.}, \bibinfo{author}{Liu, X.}, \bibinfo{author}{Xu, S.}, \bibinfo{author}{Dai, Z.}, \bibinfo{author}{Dai, Y.}, \bibinfo{author}{Xu, X.}, \bibinfo{year}{2022}.
\newblock \bibinfo{title}{Dwrseg: Rethinking efficient acquisition of multi-scale contextual information for real-time semantic segmentation}.
\newblock \bibinfo{journal}{arXiv preprint arXiv:2212.01173} .
\bibitem[{Yan-e(2011)}]{yan2011design}
\bibinfo{author}{Yan-e, D.}, \bibinfo{year}{2011}.
\newblock \bibinfo{title}{Design of intelligent agriculture management information system based on iot}, in: \bibinfo{booktitle}{2011 Fourth International Conference on Intelligent Computation Technology and Automation}, \bibinfo{organization}{IEEE}. pp. \bibinfo{pages}{1045--1049}.
\bibitem[{Yang et~al.(2023)Yang, Lei, Zhu, Cheng, Feng and Liang}]{yang2023afpn}
\bibinfo{author}{Yang, G.}, \bibinfo{author}{Lei, J.}, \bibinfo{author}{Zhu, Z.}, \bibinfo{author}{Cheng, S.}, \bibinfo{author}{Feng, Z.}, \bibinfo{author}{Liang, R.}, \bibinfo{year}{2023}.
\newblock \bibinfo{title}{Afpn: Asymptotic feature pyramid network for object detection}, in: \bibinfo{booktitle}{2023 IEEE International Conference on Systems, Man, and Cybernetics (SMC)}, \bibinfo{organization}{IEEE}. pp. \bibinfo{pages}{2184--2189}.
\bibitem[{Yu et~al.(2024)Yu, Huang, Chen, Su, Liu and Wang}]{yu2024yolo}
\bibinfo{author}{Yu, Z.}, \bibinfo{author}{Huang, H.}, \bibinfo{author}{Chen, W.}, \bibinfo{author}{Su, Y.}, \bibinfo{author}{Liu, Y.}, \bibinfo{author}{Wang, X.}, \bibinfo{year}{2024}.
\newblock \bibinfo{title}{Yolo-facev2: A scale and occlusion aware face detector}.
\newblock \bibinfo{journal}{Pattern Recognition} \bibinfo{volume}{155}, \bibinfo{pages}{110714}.

\end{thebibliography}



\end{document}